\documentclass[11pt,a4paper]{article}
\pdfoutput=1
\usepackage[hmargin=3.0cm,vmargin=3.0cm]{geometry}

\usepackage[table]{xcolor}  
\usepackage{graphicx}
\usepackage{caption,subcaption}
\usepackage{mathtools}
\usepackage{amsfonts}
\usepackage[inline]{enumitem}  
\usepackage{hyperref}
\usepackage[noabbrev]{cleveref}  
\usepackage{authblk}            
\hypersetup{
colorlinks=true,
citecolor=red
}

\usepackage[group-separator={,}]{siunitx}  
\usepackage{multicol}
\usepackage{booktabs}  
\usepackage{tikz}
\usetikzlibrary{matrix,chains,positioning,decorations.pathreplacing,arrows,calc,fit}
\usepackage{smartdiagram}
\usepackage[numbers]{natbib}  

\usepackage[space]{grffile}  

\usepackage{algorithm}
\usepackage{algpseudocode}

\usepackage[bottom]{footmisc}  

\newcommand\RR{\mathbb{R}}
\newcommand\EE{\mathop{\mathbb{E}}}

\newcommand\ZZ{\mathbf{Z}}

\newcommand\XX{\mathbf{X}}

\newcommand\manX{\mathcal{X}}

\newcommand\manZ{\mathcal{Z}}
\newcommand\manL{\mathcal{L}}

\newcommand\manH{\mathcal{H}}

\DeclareMathOperator*{\argmin}{arg\,min}

\newcommand\norm[1]{\left\lVert#1\right\rVert}

\begin{document}

\title{Exemplar-based synthesis of geology using kernel discrepancies and generative neural networks}




\author[]{Shing Chan\footnote{Corresponding author.\\ E-mail addresses: \texttt{sc41@hw.ac.uk} (Shing Chan), \texttt{a.elsheikh@hw.ac.uk} (Ahmed H. Elsheikh).} }
\author[]{Ahmed H. Elsheikh}
\affil[]{Heriot-Watt University, United Kingdom}
\affil[]{\small School of Energy, Geoscience, Infrastructure and Society}


\maketitle

\begin{abstract}
We propose a framework for synthesis of geological images based on an exemplar image (a.k.a. training image). We synthesize new realizations such that the discrepancy in the \emph{patch distribution} between the realizations and the exemplar image is minimized. Such discrepancy is quantified using a kernel method for two-sample test called maximum mean discrepancy. To enable fast synthesis, we train a generative neural network in an offline phase to sample realizations efficiently during deployment, while also providing a parametrization of the synthesis process. We assess the framework on a classical binary image representing channelized subsurface reservoirs, finding that the method reproduces the visual patterns and spatial statistics (image histogram and two-point probability functions) of the exemplar image. 
\end{abstract}

\section{Introduction} 

A challenge in subsurface flow simulations is to obtain a complete and accurate image
of subsurface properties, such as permeability and porosity, that are crucial
for accurate flow predictions. Since it is virtually impossible to obtain
direct measurements at every point of the domain under study, engineers can
only rely on indirect estimations of the subsurface properties, e.g. from
seismic images and sparse measurements obtained from wells. Traditionally, the
properties are modeled based on their two-point statistics; however, this
tends to produce images of the subsurface that are far from realistic. In
many scenarios, such as in channelized systems where the properties follow an
almost binary distribution and contain strong spatial correlations, two-point
statistics are not enough to describe the distribution of the properties.

This shortcoming led to the development of alternative algorithmic approaches to
synthesize subsuface images that can capture multipoint statistics.
These methods start from an exemplar image (also called training image in
the geology literature) that is deemed representative of the subsurface under
study, meaning that the spatial statistics in this image
is believed to be similar to that of the subsurface.
From there, a new image is synthesized by querying the exemplar image or deriving statistics from it, and
employing some form of randomness during the synthesis process to generate diverse outcomes.
These methods, although less theoretically founded, tend to produce
subsurface images that are more realistic than the traditional methods based on two-point statistics.
In~\citep{strebelle2001reservoir}, empirical
conditional probabilities are derived from the exemplar and used to
synthesize the new image each pixel at a time.
In~\citep{mariethoz2010direct}, a pixel is synthesized by simply querying the
exemplar and selecting pixels whose neighboring pixels match that of the current
synthesized domain.
In~\citep{tahmasebi2012multiple}, the synthesis is based on carefully copying and pasting
patches extracted from the exemplar.
As seen from the mentioned works, these approaches share many similarities with texture synthesis techniques in image
processing~\citep{efros1999texture,efros2001image,mariethoz2014bridges}.

A further challenge in subsurface flow simulations is the need to account for
inherent uncertainties of the simulations. For uncertainty quantification and
history matching, not only is it necessary to explore multiple plausible
solutions by performing simulations for a large number of realizations,
but also it is desirable that such exploration be smooth in the sense that
small changes in input parameters result in small changes in the output. By design, this
is not the case in most current synthesis algorithms. For
this reason, current approaches take a two-stage process: First, a dataset of
realistic realizations is synthesized using one of the many synthesis algorithms available;
thereafter,
parametrization~\citep{
sarma2008kernel,ma2011kernel,khaninezhad2012sparse1,vo2014new}
is performed using the dataset in a way that retains the realism of the
realizations while achieving a well-behaved function with respect to new
input parameters. It is then worth asking whether it is possible to achieve
parametric synthesis \emph{directly} from the exemplar.

Recent examples of parametric synthesis of geology from a single exemplar
include~\citep{mosser2017reconstruction,laloy2017efficient} where the authors
train generative adversarial networks~\citep{goodfellow2014generative} to
parametrize the geology using neural networks, obtaining very impressive
results. Generative adversarial networks leverage the representational power
of neural networks in two fronts: on one hand, the modeling power of neural
networks is leveraged for the realistic parametrization of images; on the other hand,
the discriminative power of neural networks is leveraged to learn the
statistics of the images. 
Since training these neural networks still require a dataset of
realizations, the approach
in~\citep{mosser2017reconstruction,laloy2017efficient} is to simply train the
neural networks on patches of the exemplar image; once trained, because
the neural networks used are convolutional, one can artificially increase the
dimension of the input parameter vector to synthesize larger images. However,
it remains unclear if this approach generalizes to arbitrary sizes.

\begin{figure}\centering
    \begin{tikzpicture}[scale=.9]
        \node (exemplar) at (3.5,2) {\includegraphics[width=1.75cm,height=1.75cm]{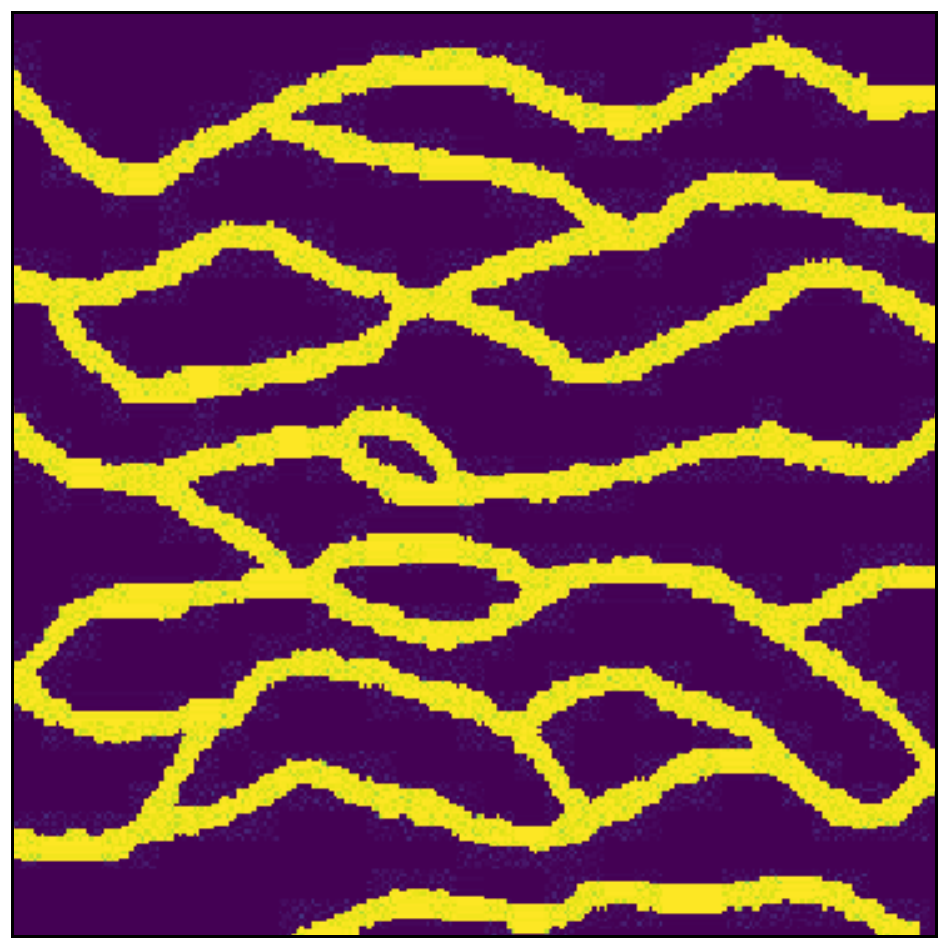}};
        \node (synthesis) [draw, minimum width=1.75cm, minimum height=1.75cm, fill=white] at (-3.5,2) {?};
        \node (mmd) at (0,-1) {$\operatorname{MMD}^2[\tilde X, \tilde X_0]$};

        \node[fill=white] (exemplar_set) at (3.5, 0)
        {$\big\{
        \raisebox{-.0cm}{\includegraphics[width=0.4cm,height=0.4cm]{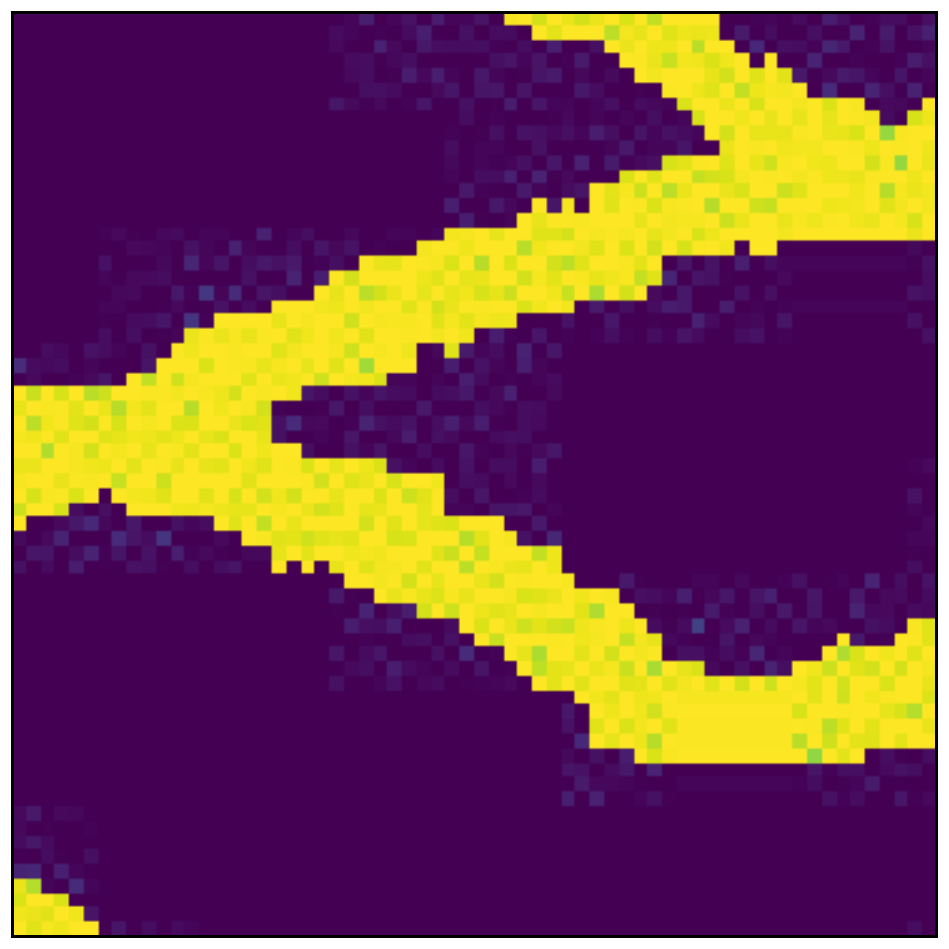}},
        \raisebox{-.0cm}{\includegraphics[width=0.4cm,height=0.4cm]{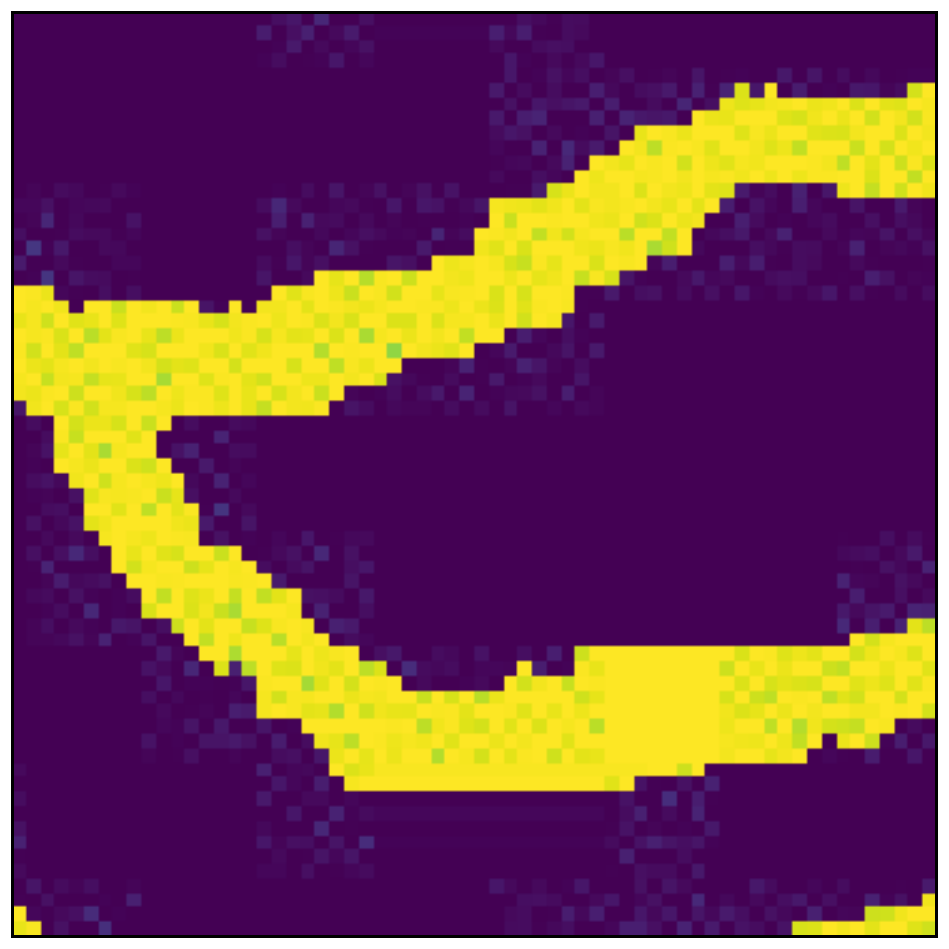}},
        \raisebox{-.0cm}{\includegraphics[width=0.4cm,height=0.4cm]{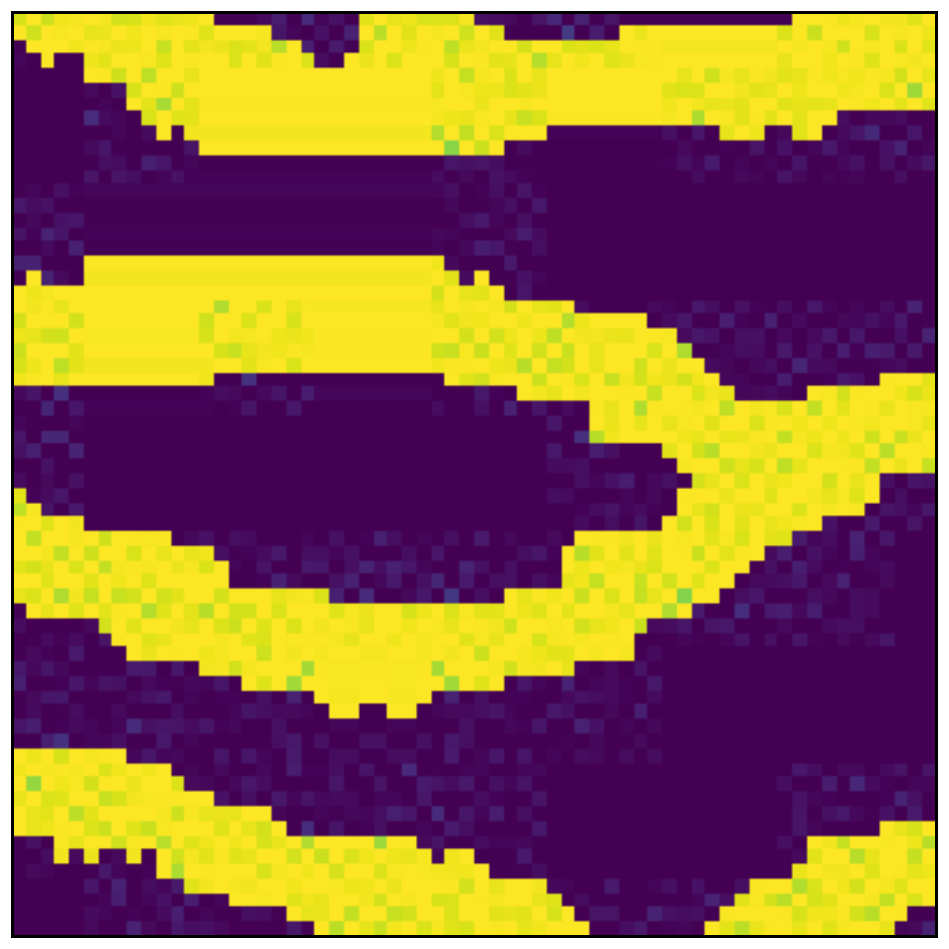}},
        \cdots
        \big\}=\tilde X_0$};

        \node[fill=white] (synthesis_set) at (-3.5, 0)
        {$\tilde X=\big\{
        \raisebox{-.0cm}{\includegraphics[width=0.4cm,height=0.4cm]{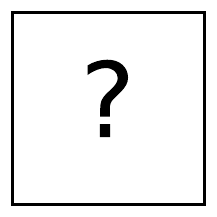}},
        \raisebox{-.0cm}{\includegraphics[width=0.4cm,height=0.4cm]{fig/square}},
        \raisebox{-.0cm}{\includegraphics[width=0.4cm,height=0.4cm]{fig/square}},
        \cdots
        \big\}$};

        \draw[->] (synthesis) -- (synthesis_set) |- (mmd);
        \draw[->] (exemplar) -- (exemplar_set) |- (mmd);

        \node (g) [left=0.5cm of synthesis] {$g_\theta(\ZZ)$};
        \node (z) [below=0.1cm of g] {$\ZZ\sim p_Z$};
        \draw [->,shorten >=3pt] (g) -- (synthesis);
    \end{tikzpicture}
    \caption{Overview of methodology. \label{fig:overview}}
\end{figure}

In this work, we propose a synthesis method to explicitly reproduce the \emph{patch distribution}
of the exemplar image. Given an exemplar image, we assume that the spatial
statistics can be sufficiently described by the distribution of
the patches extracted from the exemplar. Note that this assumption is implicit
in most exemplar-based synthesis algorithms. Then, new realizations of arbitrary
size are synthesized such that their patch distributions match that
of the exemplar. The discrepancy in distributions is measured using a kernel method for
two-sample test called \emph{maximum mean discrepancy}
(MMD)~\citep{gretton2012kernel}, then new realizations are formulated as
solutions to the minimization of this discrepancy.
To obtain a parametrization of the synthesis
process, as well as improved synthesis speed during the online phase (e.g.
for uncertainty quantification or history matching), we
assume a Markov random field model where the energy function is
(proportional to) the discrepancy,
and train a generative neural network in the offline phase to efficiently
generate new realizations online.
The resulting generator is superficially similar
to previous works~\citep{mosser2017reconstruction,laloy2017efficient}, except that we
directly synthesize the desired domain size, the statistics of the
synthesized image are explicitly compared to the exemplar, and the comparison
is done using a kernel method. An overview of the framework is shown in~\Cref{fig:overview}.

In this study, we limit ourselves to the synthesis of unconditional
realizations and leave the conditioning case for future work (examples of
conditioning to hard data can be found
in~\citep{dupont2018generating,mosser2018conditioning,chan2018parametric}). We assess our method
using the classical binary channelized image by
Strebelle~\citep{strebelle2001reservoir} of size $250\times 250$, and we
synthesize images of size $256\times 256$ and $512\times 512$.
We also study the influence of different kernels in the quality of the synthesis and
discuss alternatives for improvement.
Although not considered here, we note that the framework is
dimension-agnostic and can be directly applied to 3D images.

The rest of this work is organized as follows: 
In~\Cref{sec:background}, we describe the maximum mean discrepancy~\citep{gretton2012kernel}, and an
approach to train a generative neural network~\citep{ulyanov2017improved}.
Our main idea is presented in~\Cref{sec:methodology}: new realizations are
formulated as solutions to an optimization problem (minimize the discrepancy
in patch distributions); thereafter, a generative neural network can be
trained for fast parametric synthesis.
In~\Cref{sec:results}, we present results for the synthesis of binary channelized subsurface images based on the classical Strebelle exemplar image.
In~\Cref{sec:related}, we discuss how our framework relates to other works
and potential ideas to improve this work.
Finally, we state our conclusions and future directions in~\Cref{sec:conclusion}.
\section{Background} \label{sec:background}
\subsection{Maximum mean discrepancy}\label{sec:mmd}
Our main tool is a kernel method for two-sample test called \emph{maximum mean
discrepancy}~\citep{gretton2007kernel,gretton2012kernel}. Given two samples
$X=\{x_1,\cdots,x_m\}$ and $Y=\{y_1,\cdots,y_n\}$, the goal is to determine whether
both samples come from the same distribution. The maximum mean discrepancy (MMD)
addresses this problem by comparing the sample mean in a feature space,
\begin{align}\label{eq:mmd}
    \operatorname{MMD}^2[X,Y] &= \norm{\frac{1}{m}\sum^m_{i=1}\phi(x_i) - \frac{1}{n}\sum^n_{j=1}\phi(y_j)}^2_{\manH} \nonumber \\
    &= \frac{1}{m^2}\sum^m_{i=1}\sum^m_{i'=1}k(x_i,x_{i'}) + \frac{1}{n^2}\sum^n_{j=1}\sum^m_{j'=1}k(y_j,y_{j'})
    - \frac{2}{mn}\sum^m_{i=1}\sum^n_{j=1} k(x_i,y_j)
\end{align}
where $\phi\colon\manX\to\manH$ is a mapping to the feature space $\manH$,
and $k(x,y) \coloneqq\langle\phi(x),\phi(y)\rangle_\manH$. The useful aspect
in this formulation is that we do not need to compute
$\phi(\cdot)$ -- which can be infinite dimensional -- as long as we can
compute the function $k(\cdot,\cdot)$, called the kernel. The
kernel operator can be thought of as a similarity measure, and it must satisfy
certain properties in which case it is guaranteed to be associated to some
feature mapping/space. For an in-depth treatment,
see~\citep{gretton2012kernel}.

\paragraph{Examples} Let $x,y\in\RR^2$. For the linear kernel $k(x,y)
= x^T y$, an associated feature map is $\phi(x) = x$, then the MMD is simply the
difference in the sample mean. For the polynomial kernel of degree two
$k(x,y) = ( x^T y + 1)^2$, an associated feature map is $\phi(x) =
(x_{(2)}^2,x_{(1)}^2, \sqrt{2}x_{(2)}x_{(1)}, \sqrt{2}x_{(2)},
\sqrt{2}x_{(1)}, 1)$ where $x=(x_{(1)},x_{(2)})$, then the MMD captures differences in both mean and
covariance. Finally, the Gaussian radial basis function $k(x,y) = \exp{\{-\gamma
||x-y||^2\}}$ is associated with a mapping to infinitely many components
(obtained by Taylor expansion) and the corresponding MMD captures all the
moments of the distribution.

\subsection{Generative neural network}\label{sec:sampler}
We now describe a method to train a generative neural network as proposed in~\citep{ulyanov2017improved}. Let
$g_\theta\colon\manZ\to\manX$ be a neural network parametrized by weights
$\theta$ to be determined. The input to the neural network are realizations
of a random \emph{latent vector} $\ZZ\sim p_Z$ that provides the source of stochasticity
so that $g_\theta(\ZZ)$ is a stochastic simulator. The distribution $p_Z$ is a design
choice and is usually an easy-to-sample distribution (e.g. standard
normal distribution) so that the cost of sampling $g_\theta(\ZZ)$ is 
mostly given by the evaluation cost of $g_\theta$. Given a
target probability density function $p$, and a fixed $p_Z$, the goal is to determine $\theta$
such that $g_\theta(\ZZ)\sim p$. Let us denote the density of
$g_\theta(\ZZ)$ under $\theta$ by $q_\theta$. The Kullback-Leibler (KL) divergence
from $p$ to $q_\theta$ is
\begin{align}\label{eq:kl}
  \operatorname{D_{KL}}(q_\theta \parallel p) &= \EE_{\XX\sim q_\theta} \log \frac{q_\theta(\XX)}{p(\XX)} \\
  &= \EE_{\XX\sim q_\theta}  -\log p(\XX) + \EE_{\XX\sim q_\theta} \log q_\theta(\XX)
\end{align}
We therefore wish to find $\theta$ that minimizes this divergence. The first
term of the sum can be approximated as
\begin{equation}
\EE_{\XX\sim q_\theta}  -\log p(\XX) \approx \frac{1}{N}\sum_{i=1}^N {-\log p(X_i)}
\end{equation}
where $X_i = g_\theta(Z_i)$, by drawing $N$ realizations of $\ZZ\sim p_Z$.
The second term of the sum, called the (negative) entropy, is problematic since
we do not have the analytic form of the density $q_\theta$
($g_\theta$ normally contains multiple non-linearities). We
circumvent this issue by using a sample entropy
estimator~\citep{kozachenko1987sample,goria2005new} over a set of generated
realizations $\{X_1,\cdots,X_N\}$,
\begin{equation}\label{eq:kozachenko}
    \EE_{\XX\sim q_\theta} \log q_\theta(\XX) \approx - \hat H (\{X_i,\cdots,X_N\}) 
  \coloneqq - \frac{1}{N} \sum^N_{i=1} c \log \rho(X_i) + \mathrm{const.}
\end{equation}
where $c$ is the number of components of $\XX$, and $\rho(X_i)$ is the
distance from $X_i$ to its $k^{\mathrm{th}}$-nearest neighbor (with $k\approx
\sqrt{N}$ as a good rule of thumb~\citep{goria2005new}).
Essentially, $\hat H$ quantifies how spread the realizations are.
Putting all together, \Cref{eq:kl} can be approximated as
\begin{equation}\label{eq:kl_approx}
\operatorname{D_{KL}}(q_\theta \parallel p) \approx 
\frac{1}{N}\sum_{i=1}^N {-\log p(X_i)} - \frac{1}{N} \sum^N_{i=1} { c \log \rho(X_i) } + \mathrm{const.}
\end{equation}
Minimizing this expression can be done using gradient-based optimization,
where the gradients with respect to $\theta$ can be obtained using automatic
differentiation algorithms. Intuitively, the first term ensures that the generated samples
are in the regions of high probability of $p$, whereas the
second term ensures that the samples are diverse.

\section{Methodology} \label{sec:methodology}
We denote by $X$ an image realization and $\tilde X = \{x_1,\cdots,x_m\}$ the
corresponding set of patches extracted from $X$.
Given an exemplar image $X_0$, we assume that the spatial statistics can be
sufficiently described by the distribution estimated from the patches $\tilde X_0
= \{x^0_1,\cdots,x^0_n\}$ extracted from $X_0$. We therefore aim to
synthesize a new realization $X$ such that the patch distribution estimated
from $\tilde X = \{x_1,\cdots,x_m\}$ match the patch distribution of the exemplar. For
example, matching the distribution of ``$1\times1$ patches'' reduces to
matching the pixel histogram of the exemplar image. For patches of size $l_1
\times l_2$, the distribution to be matched is given by the multidimensional
joint histogram of $l_1l_2$ variables. The discrepancy in distributions is
measured using the maximum mean discrepancy (MMD, \Cref{sec:mmd}), then new
realizations are formulated as solutions of an optimization problem,
\begin{equation}\label{eq:optim}
    \argmin_X { \operatorname{MMD^2}[\tilde X,\tilde X_0]}
\end{equation}
with $\operatorname{MMD^2}$ defined in~\Cref{eq:mmd}.
Note that a patch $x_i$ of an image $X$ is the result of a projection
operator, therefore the minimization can be done using
gradient-based methods.
Multiple realizations can be obtained by using a local optimizer
and different initial guesses for $X$.

Optimization-based synthesis, however, can be expensive if a large number of
realizations is required in the online phase (e.g. for uncertainty
quantification or history matching); moreover, it does not provide a smooth
parametric way to explore the solution space. We therefore train a generator
in an offline phase to synthesize realizations efficiently during deployment.
We train the generator following the approach described
in~\Cref{sec:sampler}, which requires us to define a target density $p$. For this, we
shall assume a Markov random field model $p(X) \propto \exp\{-\frac{1}{\lambda}\manL(X)\}$
where $\manL(X)\coloneqq \operatorname{MMD^2}[\tilde X,\tilde X_0]$ and $\lambda$ is an unknown
``temperature'' constant (see~\citep{wu1999equivalence} or Section 4.1
of~\citep{mariethoz2014multiple} for a justification of this choice).
This conveniently sets the KL divergence in~\Cref{eq:kl} to
\begin{equation}\label{eq:kl_approx2}
\operatorname{D_{KL}}(q_\theta \parallel p) \propto
\frac{1}{N}\sum_{i=1}^N {\manL(X_i)} - \frac{\lambda}{N} \sum^N_{i=1} { c \log \rho(X_i) }
\end{equation}
where we multiplied everything by $\lambda$ and omitted the irrelevant constants. The
first term ensures that the samples minimize the MMD, while the second term
ensures that the samples are diverse. Since we do not know the constant $\lambda$, in this
work we treat it as a hyperparameter to be tuned in the offline training. In
practice, $\lambda$ acts as the trade-off between sample quality and
diversity. We summarize the steps to train the generator in~\Cref{algo:algo}.
\begin{algorithm}
  \caption{Generator training $g_\theta$}\label{algo:algo}
  \begin{algorithmic}[1]
  \Require{
    Exemplar image $X_0$, kernel $k(\cdot,\cdot)$ of MMD, ``temperature'' $\lambda$, source distribution $p_Z$, batch size $N$.
    }
    \While{$\theta$ has not converged}

      \State Sample $\{Z_1,\cdots,Z_N\} \sim p_Z$
      \State Obtain $\{X_1,\cdots,X_N\},\; X_i = g_\theta(Z_i)$
      \State $\mathbb{E} \manL \gets \frac{1}{N} \sum^N_{i=1} \operatorname{MMD^2}[\tilde X_i, \tilde X_0]$  \Comment{\Cref{eq:mmd}}
      \State $\lambda\hat{H} \gets \frac{\lambda}{N} \sum^N_{i=1} c {\log \rho(X_i)}$
      \State $\theta \gets \operatorname{Update}(\theta; \nabla_\theta{(\mathbb{E}\manL - \lambda\hat{H})})$

    \EndWhile
\end{algorithmic}
\end{algorithm}  

\subsection{Kernel choice} 

As in other kernel methods, the kernel choice is critical in the performance
of the MMD; specifically, it defines its discriminative power.
For characteristic
kernels~\citep{sriperumbudur2010hilbert,sriperumbudur2011universality}, the
MMD can distinguish two distributions in the infinite
setting~\citep{gretton2012kernel}. These include the Gaussian radial basis
function, the Laplace kernel, and the rational quadratic kernel. In this
work, we use the rational quadratic kernel $k_{\mathrm{rq}}(x,y) =
(1+\frac{||x-y||^2}{2\alpha l^2})^{-\alpha}$ due to its better gradient
behavior, where $\alpha$ and $l$ are hyperparameters to be tuned during the
offline phase (see Section 4.2 of~\citep{rasmussen2004gaussian} for
properties of this and other kernels).

Measuring similarities using kernels, however, can be challenging when the
data is very high dimensional~\citep{ramdas2015decreasing}. Moreover,
distance-based kernels (as functions of $||x-y||$) are not well-suited when
applied on the raw pixel representation of images, since differences in pixel
values are of little meaning in conveying similarity (e.g. small shifts in
pixels would imply large differences while remaining virtually the same). On
the other hand, it is often the case that the intrinsic dimensionality of the
data is low, albeit embedded in a high dimensional space. For example,
geological structures of interest such as channels can be accurately
described regardless of the grid resolution, once it is above certain
threshold. This suggests us to first project the data to a low dimensional
space, e.g. using principal component analysis or even random projections~\citep{bingham2001random},
before applying the distance-based kernel. In this work, we use the encoder
of an autoencoder trained on patches of the exemplar. The
autoencoder~\citep{hinton2006reducing} is a generalization of principal
component analysis using non-linear basis functions (represented by neural
networks). 
The idea here is to measure distances between patches using their code
representations instead of their raw pixel representations.
The resulting kernel is $k(\cdot,\cdot) = k_{\mathrm{rq}}(h(\cdot),h(\cdot))$ where $h(\cdot)$ denotes the encoder
(note that $k_{\mathrm{rq}}(h(\cdot),h(\cdot))$ \emph{is} a kernel).

\section{Numerical experiments} \label{sec:results}
\begin{figure}\centering
    \includegraphics[width=.5\textwidth]{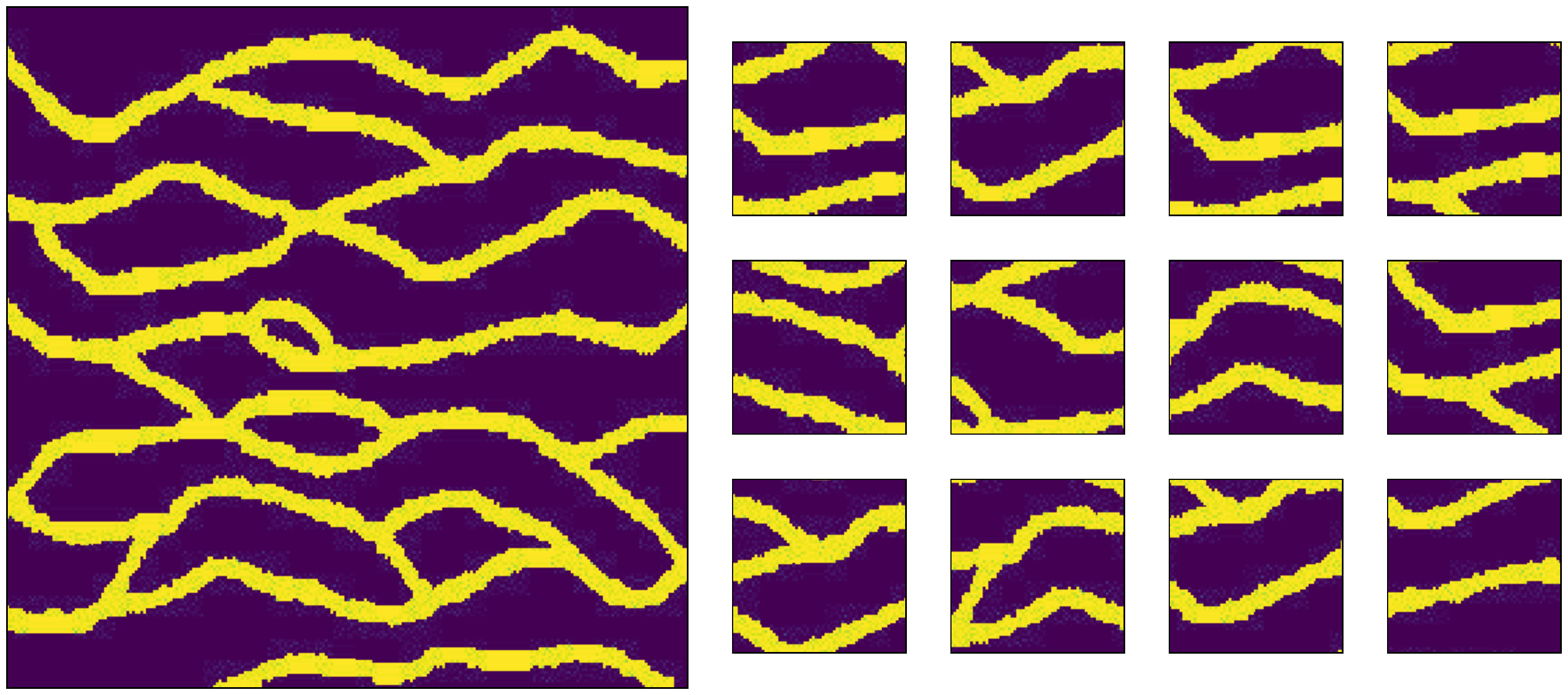}
    \caption{Exemplar image (by Strebelle~\citep{strebelle2001reservoir}) of size
    $250\times250$ depicting subsurface channels (left), and a few patches
    of size $64\times64$ extracted from the image (right). \label{fig:strebelle}}
\end{figure}

\begin{figure}\centering
    \begin{subfigure}{.9\textwidth}\centering
        \includegraphics[width=\textwidth]{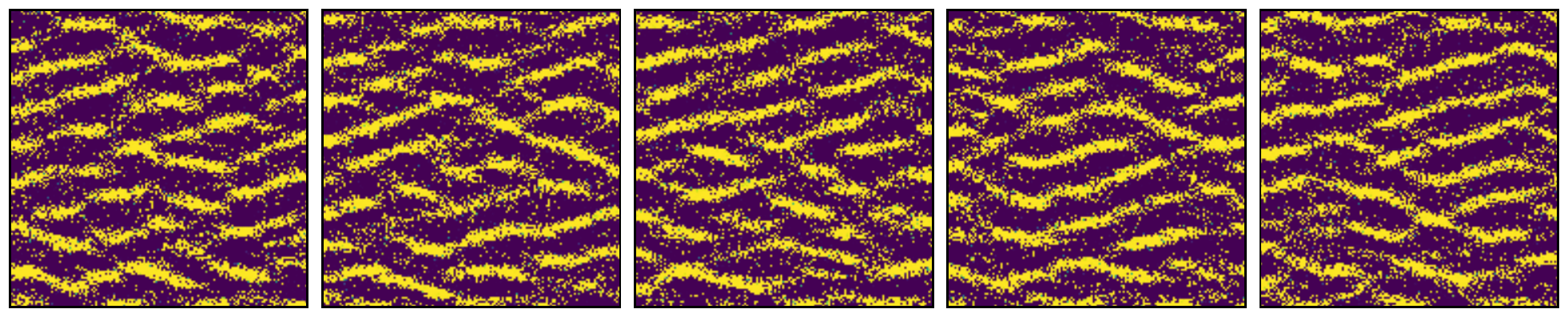}
        \caption{$k_{\mathrm{rq,randproj}}$ \label{fig:randproj}}
    \end{subfigure}\vspace{1em}

    \begin{subfigure}{.9\textwidth}\centering
        \includegraphics[width=\textwidth]{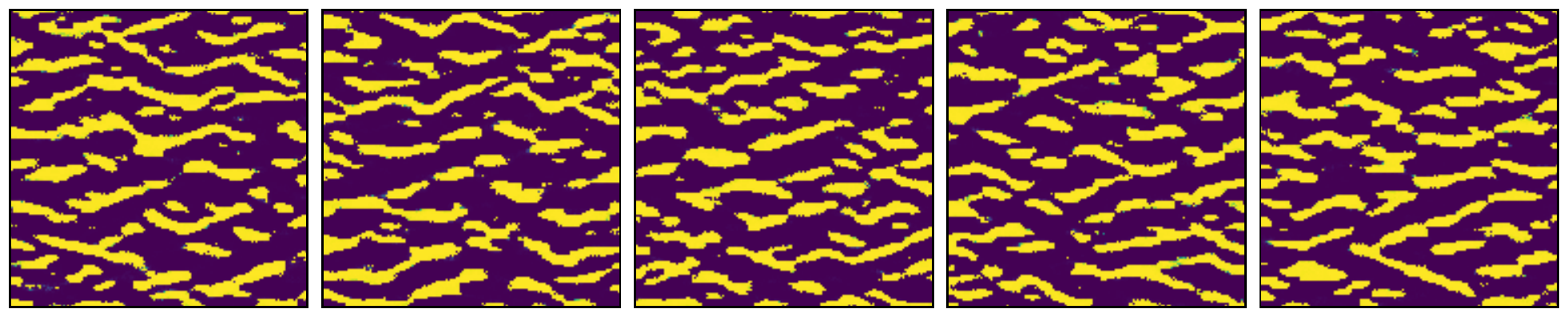}
        \caption{$k_{\mathrm{rq,pca}}$ \label{fig:pca}}
    \end{subfigure}\vspace{1em}


    \caption{Optimization-based synthesis using different kernels. 
    }
\end{figure}

We consider the synthesis of geological realizations containing subsurface
channels using the classical exemplar image of
Strebelle~\citep{strebelle2001reservoir} shown
in~\Cref{fig:strebelle}. 
Note that our target distribution is discrete (the image is binary); nevertheless,
we found good results using a continuous framework.
For convenience, we pre-process the image and work in
the $[-1,1]$ range, so that $-1$ represents the background material (blue)
and $1$ represents the channel material (yellow). The size of the exemplar image is $250\times 250$, and we use patches of size
$64\times64$. Naturally, the patch size has to be large enough to capture the
relevant patterns of interest in the exemplar image; however, it should not
be too large since this determines the amount and variability of patches
given that our exemplar is of finite size in practice.
We synthesize images of size $256\times 256$ and $512\times 512$.

For the MMD, we use a kernel of the form $k(\cdot,\cdot) =
k_{\mathrm{rq}}(h(\cdot),h(\cdot))$ where $h$ is a mapping to a lower
dimensional space, and $k_{\mathrm{rq}}$ is the rational quadratic kernel
$k_{\mathrm{rq}}(x,y) =1+\frac{||x-y||^2}{2\alpha l^2})^{-\alpha}$. We use
$\alpha=0.5$, and for $l$ (length scale parameter) we use a median
heuristic~\citep{gretton2012kernel}: we use the median distance between the patches in the
combined sample -- note that this means that our kernel adapts during the training iterations.
As for $h$, we experiment with three choices: a random projection
matrix~\citep{bingham2001random}, principal component analysis (PCA)
trained on patches of the exemplar, and the encoder of an autoencoder trained
on patches of the exemplar.

Note that the MMD in~\Cref{eq:mmd} has a quadratic cost with respect to the
sample size (although linear estimates exist~\citep{gretton2012kernel})
making it expensive to evaluate in the whole set of patches. Since we
compute the MMD iteratively, we instead evaluate on a random subset of patches
drawn during the iterations.
We draw a subset of $128$ patches. As a consequence, we found that this
procedure tends to undersample patches at the boundary of the domain, so we
perform reflection padding on the synthesis domain equal to half a patch
width before sampling patches -- this may introduce some biases at
the synthesis boundaries.

Our implementation is done using Pytorch~\citep{paszke2017automatic}, a python package for automatic
differentiation.

\begin{figure}\centering
	\begin{subfigure}{.9\textwidth}\centering
        \includegraphics[width=\textwidth]{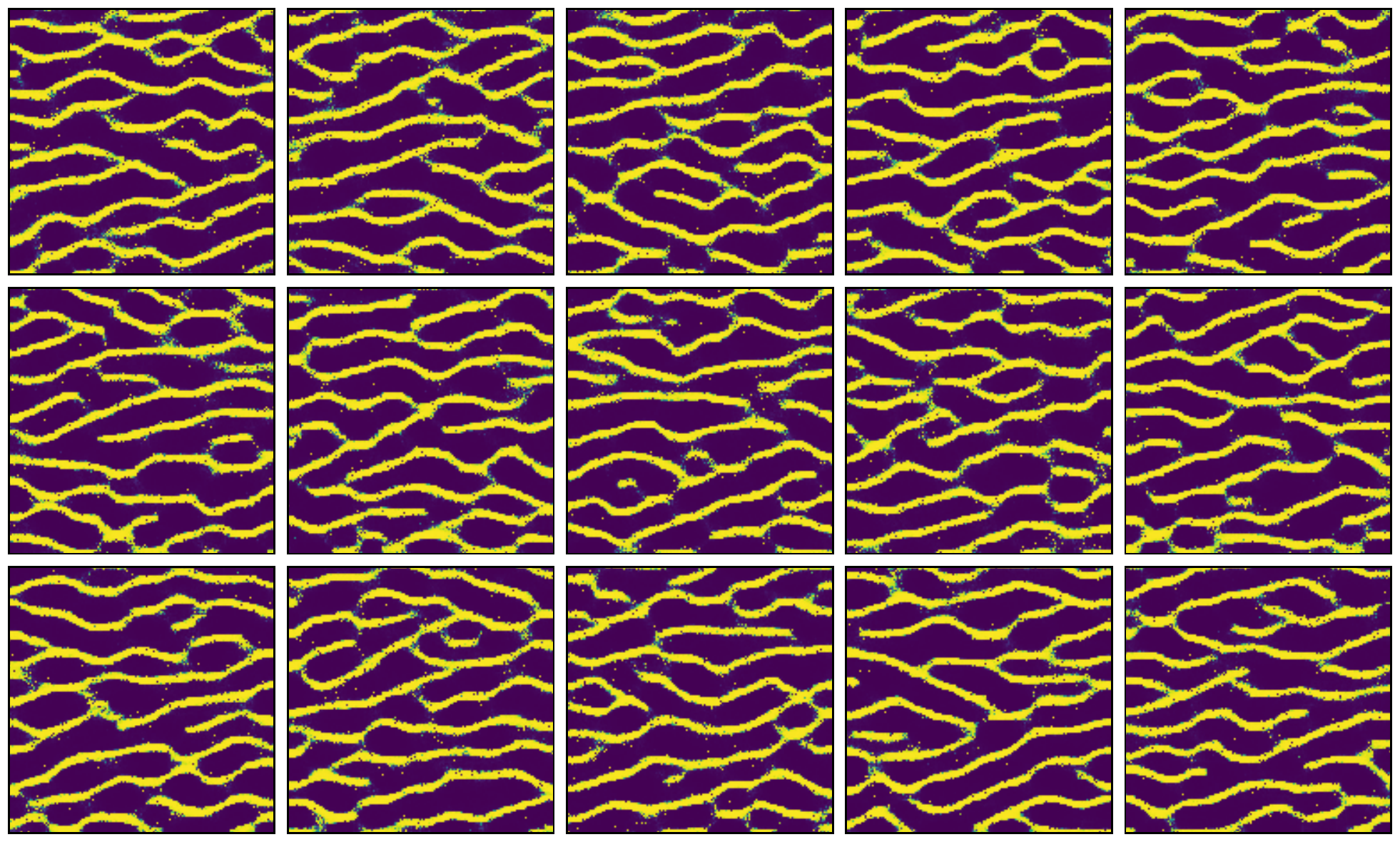}
        \caption{Random realizations ($256\times256$, optimization-based synthesis). \label{fig:optim_256}}
    \end{subfigure}\vspace{1em}

	\begin{subfigure}{.8\textwidth}\centering
        \includegraphics[width=\textwidth]{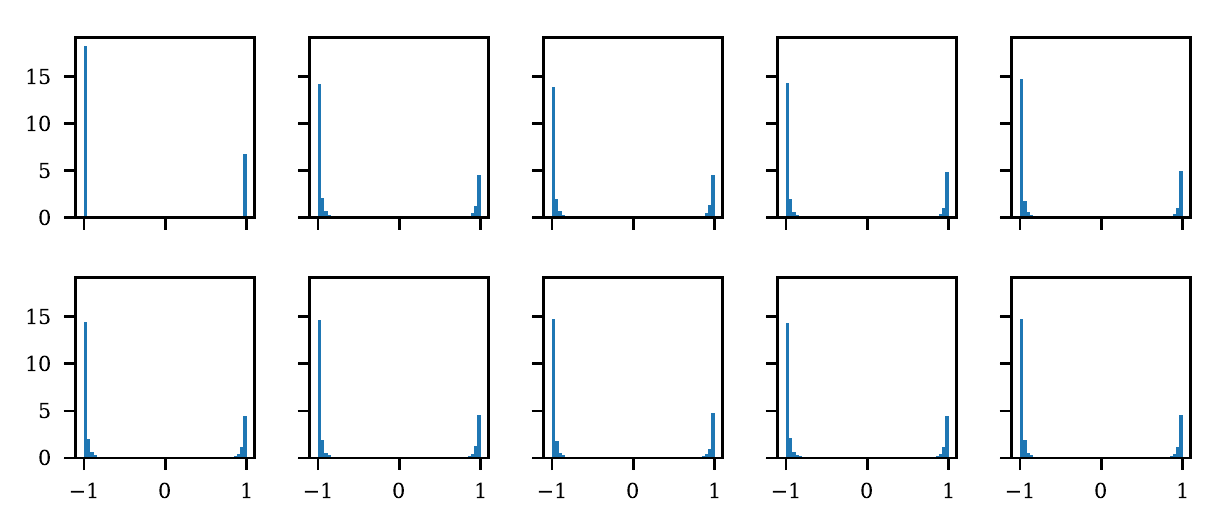}
        \caption{Image histogram of 9 random realizations. The first
        histogram (top left) corresponds to the exemplar
        image.\label{fig:optim256_hist}}
    \end{subfigure}\vspace{1em}

	\begin{subfigure}{.45\textwidth}\centering
		\includegraphics[width=\textwidth]{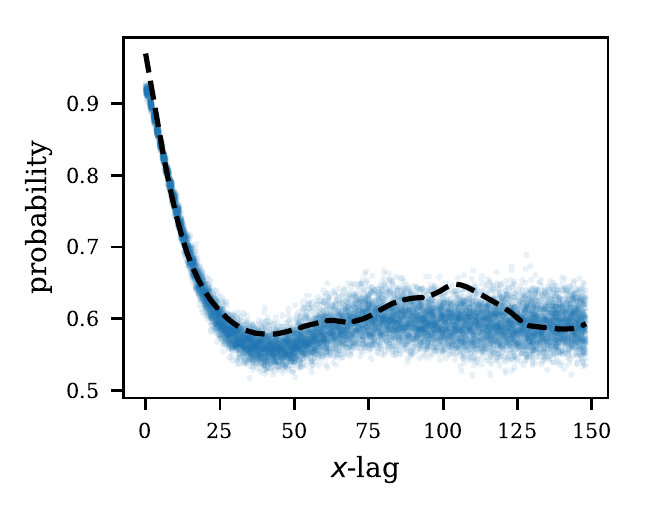}
        \caption{Two-point probability in the $x$ direction of 100 realizations. \label{fig:optim256_pfx}}
    \end{subfigure}
    \hfill
	\begin{subfigure}{.45\textwidth}\centering
        \includegraphics[width=\textwidth]{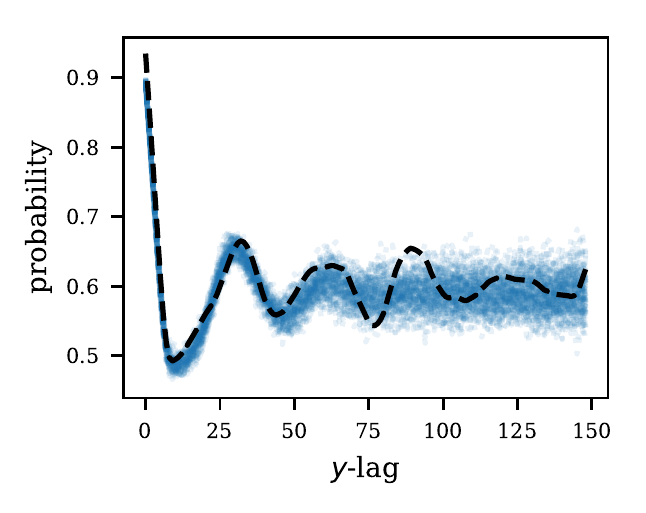}
        \caption{Two-point probability in the $y$ direction of 100 realizations. \label{fig:optim256_pfy}}
    \end{subfigure}
    \caption{Results for \emph{optimization-based} synthesis of realizations of size $256\times256$ with $k_\textrm{rq,encoder}$ kernel. \label{fig:optim256}}
\end{figure}

\subsection{Optimization-based synthesis} 

We start by synthesizing realizations using an optimization approach
(\Cref{eq:optim}). Since the pixel values are bounded in $[-1,1]$, rather
than using a constrained optimization method, here we simply reparametrize
the pixels by $X=\tanh(X')$ and solve for $X'$ instead. We use the Adam
optimizer~\citep{kingma2014adam,reddi2018convergence} (a variant of
stochastic gradient descent). We test different kernels for the MMD: For
$k_{\mathrm{rq}}$ + random projection ($k_\textrm{rq,randproj}$), we use a low-rank random matrix to
project each $64\times64$ patch to a vector of $512$ components. For
$k_{\mathrm{rq}}$ + principal component analysis (PCA) ($k_\textrm{rq,pca}$), we project each patch
to $64$ eigencomponents (retaining over 75\% of the variance). Synthesis results for
size $256\times 256$ are shown in~\Cref{fig:randproj,fig:pca}.
We can see that both random projection and PCA kernels already capture key
spatial statistics of the exemplar such as the horizontal correlations
defining the channels and the correct pixel values; however, the visual
quality of the realizations are still rather poor.

Next, we use the encoder of an autoencoder trained on the patches of the
exemplar image ($k_\textrm{rq,encoder}$). The autoencoder is trained to encode each patch into a small
code vector of size $8$, a number found via experimentation. We
experimentally found that smaller codes tend to produce better results (as
long as the autoencoder can be trained successfully), presumably by
making the distance-based kernel more accurate. Details of the autoencoder
implementation are described in~\Cref{sec:autoencoder}.

Synthesis results for size $256\times 256$ using the $k_{\mathrm{rq}}$ +
encoder kernel are summarized in~\Cref{fig:optim256}. Compared to the
previous kernels, we see that the visual quality of the realizations are
significantly improved, highlighting the impact of the kernel choice.
The synthesized images, however, still contain some spurious values such as isolated
pixels that the optimization did not manage to remove within the
iterations. If required, these could be removed using one of many available image post-processing methods~\citep{sezgin2004survey}.
We show in~\Cref{fig:optim256_hist} the normalized histogram of pixel values of nine
random realizations without thresholding, finding good correspondence with the
exemplar histogram. 
We show the two-point probability functions
(PF)~\citep{torquato1982} in the horizontal and vertical directions
in~\Cref{fig:optim256_pfx,fig:optim256_pfy}, respectively. The dashed black lines indicate the PFs of
the exemplar image, and the dotted blue lines are
PFs for $100$ random realizations. 
We do perform thresholding in this evaluation to compute the PFs.
To compute the PFs on the realizations, we randomly crop a region of size $250\times
250$ from each realization in order to match the exemplar size, and compute
the PF in this region. Note that before cropping, we first perform a
reflection padding as used in the optimization to reduce potential biases at
the boundaries.
Overall, we find good agreement between the synthesized images and
the exemplar. We show additional results for synthesis of size $512\times
512$ in~\Cref{sec:additional}.

\begin{figure}\centering
	\begin{subfigure}{.9\textwidth}\centering
        \includegraphics[width=\textwidth]{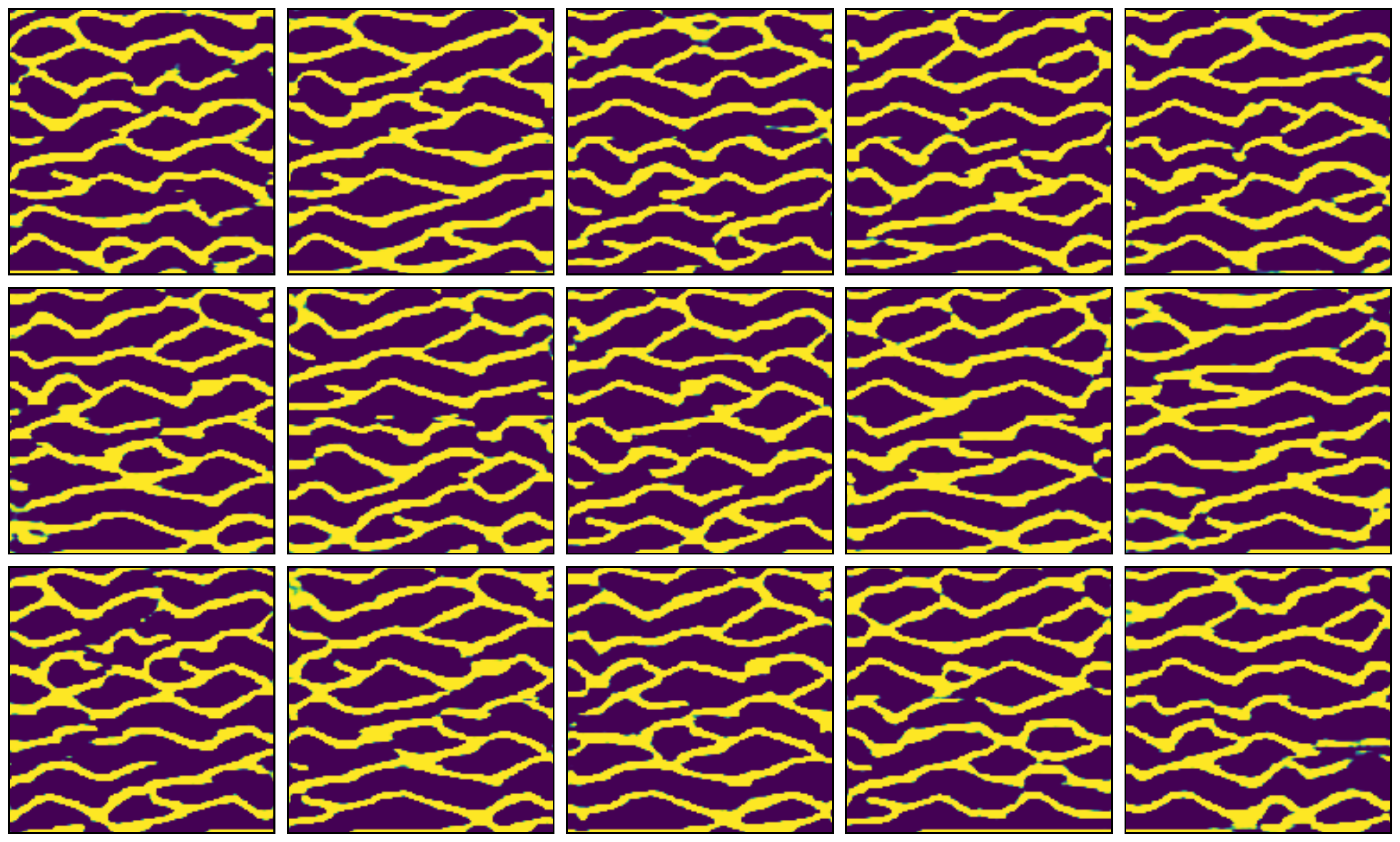}
        \caption{Random realizations ($256\times256$, generated by neural network).}
    \end{subfigure}\vspace{1em}

	\begin{subfigure}{.8\textwidth}\centering
        \includegraphics[width=\textwidth]{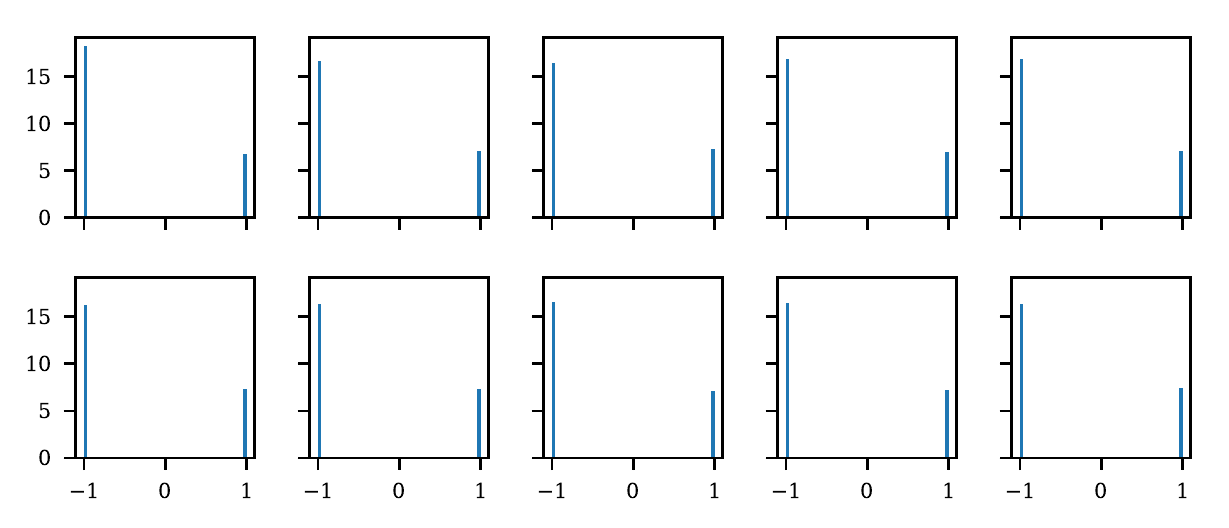}
        \caption{Image histogram of 9 random realizations. The first histogram (top left) corresponds to the exemplar image. \label{fig:nn256_hist}}
    \end{subfigure}\vspace{1em}

	\begin{subfigure}{.45\textwidth}\centering
		\includegraphics[width=\textwidth]{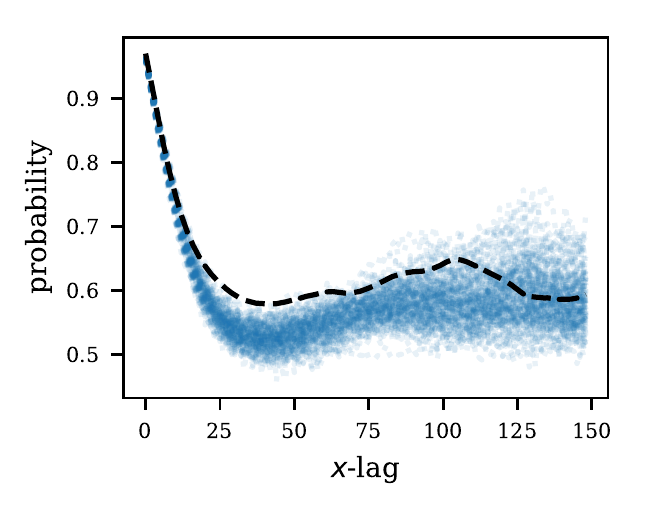}
        \caption{Two-point probability in the $x$ direction of 100 realizations. \label{fig:nn256_pfx}}
    \end{subfigure}
    \hfill
	\begin{subfigure}{.45\textwidth}\centering
        \includegraphics[width=\textwidth]{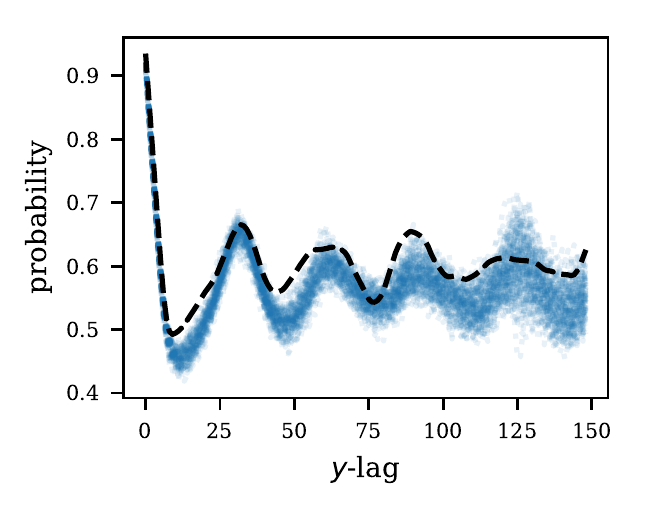}
        \caption{Two-point probability in the $y$ direction of 100 realizations. \label{fig:nn256_pfy}}
    \end{subfigure}
    \caption{Results for \emph{neural synthesis} of realizations of size $256\times256$ with $k_{\textrm{rq,encoder}}$ kernel. \label{fig:nn256}}
\end{figure}

\subsection{Neural synthesis}
We next train a generative neural network to synthesize realizations
efficiently. Here we only consider the kernel with the encoder of the autoencoder ($k_\mathrm{rq,encoder}$).
The generator is a convolutional neural network designed following the
template provided in~\citep{radford2015unsupervised}, which works well for
most computer vision tasks. Details of the architecture are given
in~\Cref{sec:generator}. To synthesize $256\times 256$ images, the generator
$g_\theta\colon \RR^{256}\to\RR^{256\times 256}$ maps from realizations of a
latent vector of size $256$ sampled from the standard normal distribution, to
image realizations of size $256\times 256$. The size of the latent vector was
chosen using a simple heuristic: proportional to the number of
non-overlapping patches in the synthesis domain times the encoding size.
We train $g_\theta$ to minimize the KL divergence in~\Cref{eq:kl_approx2},
where we use a batch size of $N=4$ and temperature hyperparameter $\lambda={10^{-8}}$.
We found that a good initial guess for $\lambda$ is a number
such that the value of the first and second terms in the KL (expected loss and
entropy, respectively) stay within the same order of magnitude in the latter
iterations of the training, so that the KL is eventually allowed to go to
zero. In fact, here we tuned $\lambda$ from $\{10^{-7}, 10^{-8}, 10^{-9}\}$,
although finer tuning is encouraged.

Results of the neural synthesis are summarized in~\Cref{fig:nn256}. Notably,
we find that the results using neural synthesis are visually better, e.g.
we do not find isolated pixels as in the optimization approach. This can be
explained by the locality prior imposed by the convolutional
architecture~\citep{saxe2011random,ulyanov2017deep}: since the image is
parametrized by a neural network, updates in the weights of the neural
network affects a whole neighborhood of the output image, in contrast to
optimization in the pixel space where pixels are updated individually;
moreover, this influence is hierarchical due to the convolutional
architecture, since layers closer to the output have a more local influence
while layers closer to the input affect the
output more globally. Regarding the normalized image histogram (again without thresholding)
in~\Cref{fig:nn256_hist}, we find that it more closely matches the true
binary shape of the exemplar histogram. Finally, we show the two-point
probability functions for the neural synthesis (computed as in the previous section)
in~\Cref{fig:nn256_pfx,fig:nn256_pfy}.
We find a slight bias in the trend of the curves, which may suggest that
further tuning of the neural network is necessary. Nonetheless, the results
remain close in relative value.

We additionally train a generator $g_\theta\colon\RR^{512}\to\RR^{512\times
512}$ to synthesize realizations of size $512\times 512$. For this case, we
use a latent vector of size $512$ (also with standard normal distribution) and $\lambda=10^{-9}$.
The results are summarized in~\Cref{sec:additional}.

\begin{figure}\centering
    \includegraphics[width=.9\textwidth]{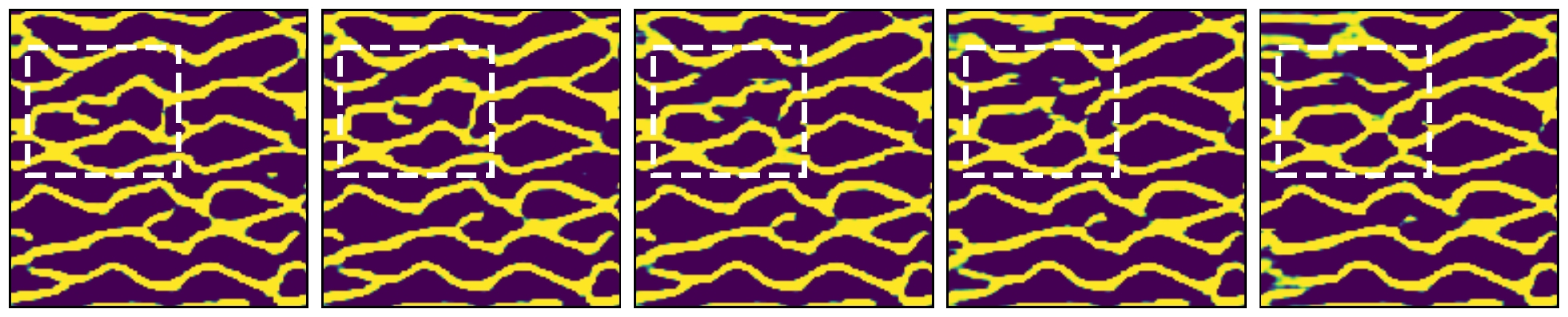}
    \caption{Linear interpolation of one coordinate of the latent vector. \label{fig:smooth}}
\end{figure}

\paragraph{Smooth transitions}
Since $g_\theta$ is continuous by construction, small changes in the input results in
small changes in the output. We verify this in~\Cref{fig:smooth} where we
show the outputs of the generator of size $256\times 256$. Starting from an
initial random realization of the latent vector $\ZZ$, we linearly vary one
of its coordinates while fixing the remaining coordinates.
Note that unlike methods such as principal component analysis where the latent vector
represents the coefficients of the eigenvectors, the latent vector of
generative neural networks lack interpretability. Learning interpretable
latent vectors is an ongoing area of research,
see e.g.~\citep{chen2016infogan}.
\section{Related work} \label{sec:related}

\begin{figure}\centering
    \begin{subfigure}{.9\textwidth}\centering
        \includegraphics[width=\textwidth]{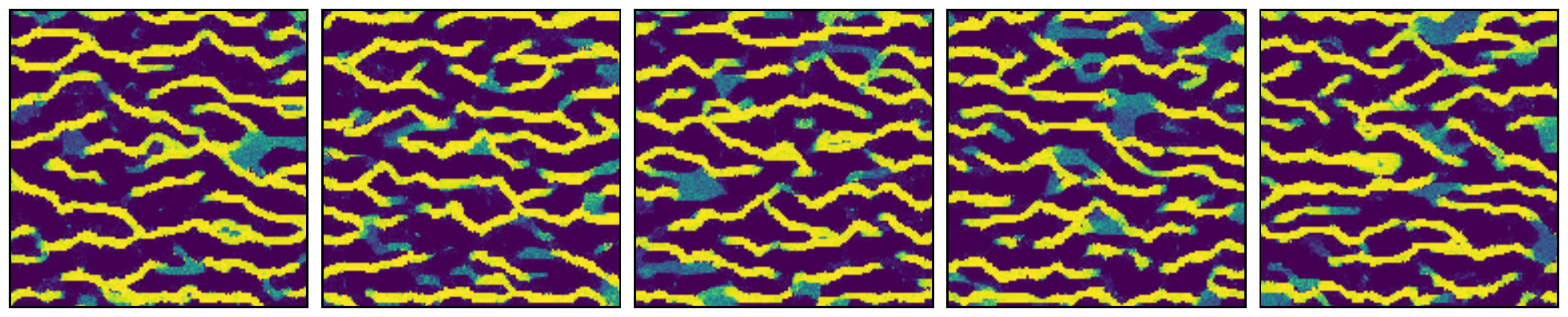}
        \caption{$k_{\mathrm{poly,vgg}}$ \label{fig:vgg}}
    \end{subfigure}\vspace{1em}

    \begin{subfigure}{.9\textwidth}\centering
        \includegraphics[width=\textwidth]{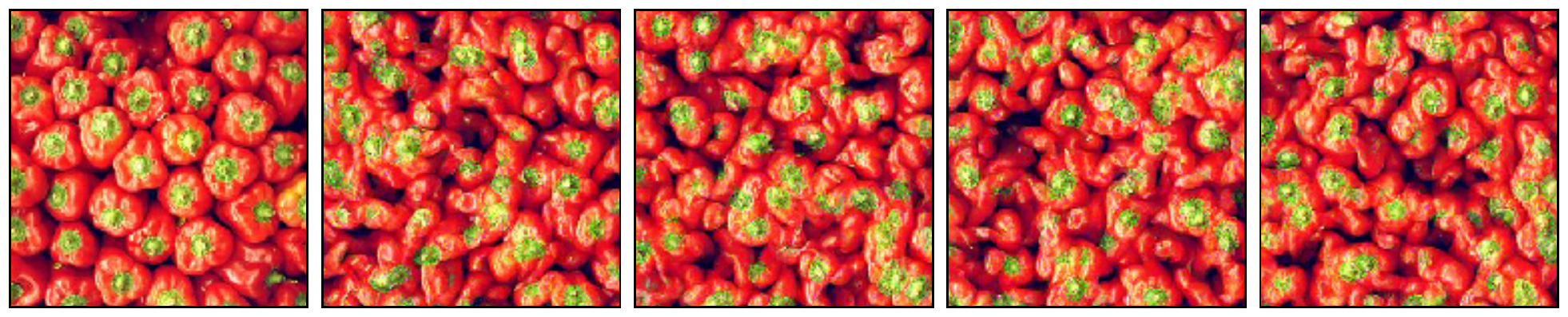}
        \caption{$k_{\mathrm{poly,vgg}}$ \label{fig:vgg_peppers}}
    \end{subfigure}
    \caption{Optimization-based synthesis using the kernel
    from~\citep{gatys2015neural}, i.e. VGG encoder + polynomial kernel of
    second degree. Compare (a) with \Cref{fig:randproj}, \Cref{fig:pca}, and~\Cref{fig:optim_256}.}
\end{figure}

\paragraph{Neural kernels} The seminal work
in~\citep{gatys2015neural} showed that it is possible to synthesize textures
from an exemplar by matching statistics of feature responses of a
pre-trained neural network evaluated on the exemplar. Briefly, the exemplar is fed into the
VGG-net~\citep{simonyan2014very} -- a very large neural network trained on
natural images for classification -- and a matrix is formed containing the
correlations of feature responses at layers of the neural
network. Then, new realizations are synthesized such that their corresponding
matrices are close to that of the exemplar. It was later shown
in~\citep{li2017demystifying} that this is equivalent to computing the
maximum mean discrepancy on the feature responses using the polynomial kernel
$k(x,y)=(x^T y)^2$. 
Finally, by noting that
each feature response corresponds to a patch of the domain (defined by its
receptive field), we conclude that this is an instance of our framework where the kernel 
is composed of a polynomial kernel and
the VGG-net as ``encoder''. Note that in this case, the encoder is trained on
a different task (classification) using large sets of other images, making the approach
an example of transfer learning~\citep{pratt1993discriminability,pan2010survey}. We show synthesis results
using this kernel in~\Cref{fig:vgg,fig:vgg_peppers} for our geological image
and for a natural texture (peppers; first image in the row), respectively. We
see that the kernel performs very well for the image of peppers, but not so well
for our binary geological image -- presumably because the VGG-net is
trained on natural color images.

\paragraph{Neural generators} The present approach to train a generator is
based on~\citep{ulyanov2017improved} where a sample entropy estimator based
on the nearest neighbor is used. Here we use a $k^{\mathrm{th}}$ nearest
neighbor~\citep{goria2005new} estimation which we found to be numerically
more stable. These estimators, however, measure the distance between
realizations in the raw representation, which for images may not be well
suited. An ad-hoc alternative can be found in~\citep{li2017diversified} where
distances are instead computed on the feature responses of a neural
network evaluated on the images. Other alternatives include normalizing
flow~\citep{rezende2015variational}, autoregressive
flow~\citep{kingma2016improved}, and Stein variational gradient
descent~\citep{feng2017learning}. The latter is an interesting alternative
which involves yet another kernel to estimate the average diversity in the
sample, making it useful for embedding prior knowledge about the geology. 

\paragraph{Adaptive kernels} The kernel has a big influence on the quality of
the synthesis since it defines the discriminative power of the MMD. In this
work, we first reduce the data using a fixed encoder of an autoencoder
previously trained on patches of the exemplar image. The same approach is
employed in~\citep{li2015generative} in the context of generative modeling of
natural images~\citep{dziugaite2015training}. 
This is done following the intuition that distances in the code representation of
an autoencoder are more suitable than in the raw pixel representation of images and spatial data.
This manual kernel engineering
can be circumvented by considering adaptive kernels~\citep{li2017mmd}. 
The idea here is to iteratively update the kernel encoder during the training
iterations, thus serving as an adversary maximizing the MMD whereas the
generator is trained to minimize it.
This idea can be taken further by
considering other functions aside from kernels, e.g. parametrized by neural networks~\citep{goodfellow2014generative}.
All these methods involve adversarial training of an additional neural
network as well as dynamic target loss functions, involving
numerical challenges in terms of stability as well as computational cost.
On the upside, they tend to produce state-of-the-art results in computer vision.

\section{Conclusion} \label{sec:conclusion}
We introduced a synthesis method for geological images consisting of
minimizing the discrepancy in the patch distribution between an exemplar image and the synthesized image. To make the
synthesis parametric and efficient, a neural network is trained in an offline
phase to sample realizations quickly during deployment. We assessed the
framework using the classical exemplar image by Strebelle
of size $250\times 250$, and synthesize images of sizes $256\times 256$ and $512\times 512$.
We find that with an adequate kernel, the visual patterns from the exemplar image are clearly
reproduced, and the spatial statistics as measured by the image histogram and
the two-point probability functions show good agreement with respect to the
exemplar.
Our framework depends on the discriminative power of the MMD, which is highly
influenced by the kernel choice, as verified in our work when
synthesizing using different kernels. 
To train the generative neural network, we currently use a sample entropy
estimator based on distances in pixel space, which might not be ideal for spatial data.
We discussed possible improvements to our framework such as
adaptive kernels and kernel-based training of the generator. These are
worth exploring in future work, as well as incorporating soft and hard
conditioning.
\section*{Acknowledgements}
We would like to thank Yanghao Li, Dmitry Ulyanov, and Yijun Li for their helpful comments.
\bibliographystyle{plainnat}
\bibliography{biblio}

\begin{thebibliography}{51}
\providecommand{\natexlab}[1]{#1}
\providecommand{\url}[1]{\texttt{#1}}
\expandafter\ifx\csname urlstyle\endcsname\relax
  \providecommand{\doi}[1]{doi: #1}\else
  \providecommand{\doi}{doi: \begingroup \urlstyle{rm}\Url}\fi

\bibitem[Bingham and Mannila(2001)]{bingham2001random}
Ella Bingham and Heikki Mannila.
\newblock Random projection in dimensionality reduction: applications to image
  and text data.
\newblock In \emph{Proceedings of the seventh ACM SIGKDD international
  conference on Knowledge discovery and data mining}, pages 245--250. ACM,
  2001.

\bibitem[Chan and Elsheikh(2018)]{chan2018parametric}
Shing Chan and Ahmed~H Elsheikh.
\newblock Parametric generation of conditional geological realizations using
  generative neural networks.
\newblock \emph{arXiv preprint arXiv:1807.05207}, 2018.

\bibitem[Chen et~al.(2016)Chen, Duan, Houthooft, Schulman, Sutskever, and
  Abbeel]{chen2016infogan}
Xi~Chen, Yan Duan, Rein Houthooft, John Schulman, Ilya Sutskever, and Pieter
  Abbeel.
\newblock Infogan: Interpretable representation learning by information
  maximizing generative adversarial nets.
\newblock In \emph{Advances in neural information processing systems}, pages
  2172--2180, 2016.

\bibitem[Dupont et~al.(2018)Dupont, Zhang, Tilke, Liang, and
  Bailey]{dupont2018generating}
Emilien Dupont, Tuanfeng Zhang, Peter Tilke, Lin Liang, and William Bailey.
\newblock Generating realistic geology conditioned on physical measurements
  with generative adversarial networks.
\newblock \emph{arXiv preprint arXiv:1802.03065}, 2018.

\bibitem[Dziugaite et~al.(2015)Dziugaite, Roy, and
  Ghahramani]{dziugaite2015training}
Gintare~Karolina Dziugaite, Daniel~M Roy, and Zoubin Ghahramani.
\newblock Training generative neural networks via maximum mean discrepancy
  optimization.
\newblock \emph{arXiv preprint arXiv:1505.03906}, 2015.

\bibitem[Efros and Freeman(2001)]{efros2001image}
Alexei~A Efros and William~T Freeman.
\newblock Image quilting for texture synthesis and transfer.
\newblock In \emph{Proceedings of the 28th annual conference on Computer
  graphics and interactive techniques}, pages 341--346. ACM, 2001.

\bibitem[Efros and Leung(1999)]{efros1999texture}
Alexei~A Efros and Thomas~K Leung.
\newblock Texture synthesis by non-parametric sampling.
\newblock In \emph{iccv}, page 1033. IEEE, 1999.

\bibitem[Feng et~al.(2017)Feng, Wang, and Liu]{feng2017learning}
Yihao Feng, Dilin Wang, and Qiang Liu.
\newblock Learning to draw samples with amortized stein variational gradient
  descent.
\newblock \emph{arXiv preprint arXiv:1707.06626}, 2017.

\bibitem[Gatys et~al.(2015)Gatys, Ecker, and Bethge]{gatys2015neural}
Leon~A Gatys, Alexander~S Ecker, and Matthias Bethge.
\newblock A neural algorithm of artistic style.
\newblock \emph{arXiv preprint arXiv:1508.06576}, 2015.

\bibitem[Goodfellow et~al.(2014)Goodfellow, Pouget-Abadie, Mirza, Xu,
  Warde-Farley, Ozair, Courville, and Bengio]{goodfellow2014generative}
Ian Goodfellow, Jean Pouget-Abadie, Mehdi Mirza, Bing Xu, David Warde-Farley,
  Sherjil Ozair, Aaron Courville, and Yoshua Bengio.
\newblock Generative adversarial nets.
\newblock In \emph{Advances in neural information processing systems}, pages
  2672--2680, 2014.

\bibitem[Goria et~al.(2005)Goria, Leonenko, Mergel, and
  Novi~Inverardi]{goria2005new}
Mohammed~Nawaz Goria, Nikolai~N Leonenko, Victor~V Mergel, and Pier~Luigi
  Novi~Inverardi.
\newblock A new class of random vector entropy estimators and its applications
  in testing statistical hypotheses.
\newblock \emph{Journal of Nonparametric Statistics}, 17\penalty0 (3):\penalty0
  277--297, 2005.

\bibitem[Gretton et~al.(2007)Gretton, Borgwardt, Rasch, Sch{\"o}lkopf, and
  Smola]{gretton2007kernel}
Arthur Gretton, Karsten~M Borgwardt, Malte Rasch, Bernhard Sch{\"o}lkopf, and
  Alex~J Smola.
\newblock A kernel method for the two-sample-problem.
\newblock In \emph{Advances in neural information processing systems}, pages
  513--520, 2007.

\bibitem[Gretton et~al.(2012)Gretton, Borgwardt, Rasch, Sch{\"o}lkopf, and
  Smola]{gretton2012kernel}
Arthur Gretton, Karsten~M Borgwardt, Malte~J Rasch, Bernhard Sch{\"o}lkopf, and
  Alexander Smola.
\newblock A kernel two-sample test.
\newblock \emph{Journal of Machine Learning Research}, 13\penalty0
  (Mar):\penalty0 723--773, 2012.

\bibitem[Hinton and Salakhutdinov(2006)]{hinton2006reducing}
Geoffrey~E Hinton and Ruslan~R Salakhutdinov.
\newblock Reducing the dimensionality of data with neural networks.
\newblock \emph{science}, 313\penalty0 (5786):\penalty0 504--507, 2006.

\bibitem[Khaninezhad et~al.(2012)Khaninezhad, Jafarpour, and
  Li]{khaninezhad2012sparse1}
Mohammadreza~Mohammad Khaninezhad, Behnam Jafarpour, and Lianlin Li.
\newblock Sparse geologic dictionaries for subsurface flow model calibration:
  Part i. inversion formulation.
\newblock \emph{Advances in Water Resources}, 39:\penalty0 106--121, 2012.

\bibitem[Kingma and Ba(2014)]{kingma2014adam}
Diederik Kingma and Jimmy Ba.
\newblock Adam: A method for stochastic optimization.
\newblock \emph{arXiv preprint arXiv:1412.6980}, 2014.

\bibitem[Kingma et~al.(2016)Kingma, Salimans, Jozefowicz, Chen, Sutskever, and
  Welling]{kingma2016improved}
Diederik~P Kingma, Tim Salimans, Rafal Jozefowicz, Xi~Chen, Ilya Sutskever, and
  Max Welling.
\newblock Improved variational inference with inverse autoregressive flow.
\newblock In \emph{Advances in Neural Information Processing Systems}, pages
  4743--4751, 2016.

\bibitem[Kozachenko and Leonenko(1987)]{kozachenko1987sample}
LF~Kozachenko and Nikolai~N Leonenko.
\newblock Sample estimate of the entropy of a random vector.
\newblock \emph{Problemy Peredachi Informatsii}, 23\penalty0 (2):\penalty0
  9--16, 1987.

\bibitem[Laloy et~al.(2017)Laloy, H{\'e}rault, Jacques, and
  Linde]{laloy2017efficient}
Eric Laloy, Romain H{\'e}rault, Diederik Jacques, and Niklas Linde.
\newblock Efficient training-image based geostatistical simulation and
  inversion using a spatial generative adversarial neural network.
\newblock \emph{arXiv preprint arXiv:1708.04975}, 2017.

\bibitem[Li et~al.(2017{\natexlab{a}})Li, Chang, Cheng, Yang, and
  P{\'o}czos]{li2017mmd}
Chun-Liang Li, Wei-Cheng Chang, Yu~Cheng, Yiming Yang, and Barnab{\'a}s
  P{\'o}czos.
\newblock Mmd gan: Towards deeper understanding of moment matching network.
\newblock In \emph{Advances in Neural Information Processing Systems}, pages
  2203--2213, 2017{\natexlab{a}}.

\bibitem[Li et~al.(2017{\natexlab{b}})Li, Wang, Liu, and
  Hou]{li2017demystifying}
Yanghao Li, Naiyan Wang, Jiaying Liu, and Xiaodi Hou.
\newblock Demystifying neural style transfer.
\newblock \emph{arXiv preprint arXiv:1701.01036}, 2017{\natexlab{b}}.

\bibitem[Li et~al.(2017{\natexlab{c}})Li, Fang, Yang, Wang, Lu, and
  Yang]{li2017diversified}
Yijun Li, Chen Fang, Jimei Yang, Zhaowen Wang, Xin Lu, and Ming-Hsuan Yang.
\newblock Diversified texture synthesis with feed-forward networks.
\newblock In \emph{Proc. CVPR}, 2017{\natexlab{c}}.

\bibitem[Li et~al.(2015)Li, Swersky, and Zemel]{li2015generative}
Yujia Li, Kevin Swersky, and Rich Zemel.
\newblock Generative moment matching networks.
\newblock In \emph{International Conference on Machine Learning}, pages
  1718--1727, 2015.

\bibitem[Ma and Zabaras(2011)]{ma2011kernel}
Xiang Ma and Nicholas Zabaras.
\newblock Kernel principal component analysis for stochastic input model
  generation.
\newblock \emph{Journal of Computational Physics}, 230\penalty0 (19):\penalty0
  7311--7331, 2011.

\bibitem[Mariethoz and Caers(2014)]{mariethoz2014multiple}
Gregoire Mariethoz and Jef Caers.
\newblock \emph{Multiple-point geostatistics: stochastic modeling with training
  images}.
\newblock John Wiley \& Sons, 2014.

\bibitem[Mariethoz and Lefebvre(2014)]{mariethoz2014bridges}
Gregoire Mariethoz and Sylvain Lefebvre.
\newblock Bridges between multiple-point geostatistics and texture synthesis:
  Review and guidelines for future research.
\newblock \emph{Computers \& Geosciences}, 66:\penalty0 66--80, 2014.

\bibitem[Mariethoz et~al.(2010)Mariethoz, Renard, and
  Straubhaar]{mariethoz2010direct}
Gregoire Mariethoz, Philippe Renard, and Julien Straubhaar.
\newblock The direct sampling method to perform multiple-point geostatistical
  simulations.
\newblock \emph{Water Resources Research}, 46\penalty0 (11), 2010.

\bibitem[Mosser et~al.(2017)Mosser, Dubrule, and
  Blunt]{mosser2017reconstruction}
Lukas Mosser, Olivier Dubrule, and Martin~J Blunt.
\newblock Reconstruction of three-dimensional porous media using generative
  adversarial neural networks.
\newblock \emph{arXiv preprint arXiv:1704.03225}, 2017.

\bibitem[Mosser et~al.(2018)Mosser, Dubrule, and Blunt]{mosser2018conditioning}
Lukas Mosser, Olivier Dubrule, and Martin~J Blunt.
\newblock Conditioning of three-dimensional generative adversarial networks for
  pore and reservoir-scale models.
\newblock \emph{arXiv preprint arXiv:1802.05622}, 2018.

\bibitem[Odena et~al.(2016)Odena, Dumoulin, and Olah]{odena2016deconvolution}
Augustus Odena, Vincent Dumoulin, and Chris Olah.
\newblock Deconvolution and checkerboard artifacts.
\newblock \emph{Distill}, 2016.
\newblock \doi{10.23915/distill.00003}.
\newblock URL \url{http://distill.pub/2016/deconv-checkerboard}.

\bibitem[Pan and Yang(2010)]{pan2010survey}
Sinno~Jialin Pan and Qiang Yang.
\newblock A survey on transfer learning.
\newblock \emph{IEEE Transactions on knowledge and data engineering},
  22\penalty0 (10):\penalty0 1345--1359, 2010.

\bibitem[Paszke et~al.(2017)Paszke, Gross, Chintala, Chanan, Yang, DeVito, Lin,
  Desmaison, Antiga, and Lerer]{paszke2017automatic}
Adam Paszke, Sam Gross, Soumith Chintala, Gregory Chanan, Edward Yang, Zachary
  DeVito, Zeming Lin, Alban Desmaison, Luca Antiga, and Adam Lerer.
\newblock Automatic differentiation in pytorch.
\newblock 2017.

\bibitem[Pratt(1993)]{pratt1993discriminability}
Lorien~Y Pratt.
\newblock Discriminability-based transfer between neural networks.
\newblock In \emph{Advances in neural information processing systems}, pages
  204--211, 1993.

\bibitem[Radford et~al.(2015)Radford, Metz, and
  Chintala]{radford2015unsupervised}
Alec Radford, Luke Metz, and Soumith Chintala.
\newblock Unsupervised representation learning with deep convolutional
  generative adversarial networks.
\newblock \emph{arXiv preprint arXiv:1511.06434}, 2015.

\bibitem[Ramdas et~al.(2015)Ramdas, Reddi, P{\'o}czos, Singh, and
  Wasserman]{ramdas2015decreasing}
Aaditya Ramdas, Sashank~Jakkam Reddi, Barnab{\'a}s P{\'o}czos, Aarti Singh, and
  Larry~A Wasserman.
\newblock On the decreasing power of kernel and distance based nonparametric
  hypothesis tests in high dimensions.
\newblock In \emph{AAAI}, pages 3571--3577, 2015.

\bibitem[Rasmussen(2004)]{rasmussen2004gaussian}
Carl~Edward Rasmussen.
\newblock Gaussian processes in machine learning.
\newblock In \emph{Advanced lectures on machine learning}, pages 63--71.
  Springer, 2004.

\bibitem[Reddi et~al.(2018)Reddi, Kale, and Kumar]{reddi2018convergence}
Sashank~J Reddi, Satyen Kale, and Sanjiv Kumar.
\newblock On the convergence of adam and beyond.
\newblock 2018.

\bibitem[Rezende and Mohamed(2015)]{rezende2015variational}
Danilo~Jimenez Rezende and Shakir Mohamed.
\newblock Variational inference with normalizing flows.
\newblock \emph{arXiv preprint arXiv:1505.05770}, 2015.

\bibitem[Sarma et~al.(2008)Sarma, Durlofsky, and Aziz]{sarma2008kernel}
Pallav Sarma, Louis~J Durlofsky, and Khalid Aziz.
\newblock Kernel principal component analysis for efficient, differentiable
  parameterization of multipoint geostatistics.
\newblock \emph{Mathematical Geosciences}, 40\penalty0 (1):\penalty0 3--32,
  2008.

\bibitem[Saxe et~al.(2011)Saxe, Koh, Chen, Bhand, Suresh, and
  Ng]{saxe2011random}
Andrew~M Saxe, Pang~Wei Koh, Zhenghao Chen, Maneesh Bhand, Bipin Suresh, and
  Andrew~Y Ng.
\newblock On random weights and unsupervised feature learning.
\newblock In \emph{ICML}, pages 1089--1096, 2011.

\bibitem[Sezgin and Sankur(2004)]{sezgin2004survey}
Mehmet Sezgin and B{\"u}lent Sankur.
\newblock Survey over image thresholding techniques and quantitative
  performance evaluation.
\newblock \emph{Journal of Electronic imaging}, 13\penalty0 (1):\penalty0
  146--166, 2004.

\bibitem[Simonyan and Zisserman(2014)]{simonyan2014very}
Karen Simonyan and Andrew Zisserman.
\newblock Very deep convolutional networks for large-scale image recognition.
\newblock \emph{arXiv preprint arXiv:1409.1556}, 2014.

\bibitem[Sriperumbudur et~al.(2010)Sriperumbudur, Gretton, Fukumizu,
  Sch{\"o}lkopf, and Lanckriet]{sriperumbudur2010hilbert}
Bharath~K Sriperumbudur, Arthur Gretton, Kenji Fukumizu, Bernhard
  Sch{\"o}lkopf, and Gert~RG Lanckriet.
\newblock Hilbert space embeddings and metrics on probability measures.
\newblock \emph{Journal of Machine Learning Research}, 11\penalty0
  (Apr):\penalty0 1517--1561, 2010.

\bibitem[Sriperumbudur et~al.(2011)Sriperumbudur, Fukumizu, and
  Lanckriet]{sriperumbudur2011universality}
Bharath~K Sriperumbudur, Kenji Fukumizu, and Gert~RG Lanckriet.
\newblock Universality, characteristic kernels and rkhs embedding of measures.
\newblock \emph{Journal of Machine Learning Research}, 12\penalty0
  (Jul):\penalty0 2389--2410, 2011.

\bibitem[Strebelle and Journel(2001)]{strebelle2001reservoir}
Sebastien~B Strebelle and Andre~G Journel.
\newblock Reservoir modeling using multiple-point statistics.
\newblock In \emph{SPE Annual Technical Conference and Exhibition}. Society of
  Petroleum Engineers, 2001.

\bibitem[Tahmasebi et~al.(2012)Tahmasebi, Hezarkhani, and
  Sahimi]{tahmasebi2012multiple}
Pejman Tahmasebi, Ardeshir Hezarkhani, and Muhammad Sahimi.
\newblock Multiple-point geostatistical modeling based on the cross-correlation
  functions.
\newblock \emph{Computational Geosciences}, 16\penalty0 (3):\penalty0 779--797,
  2012.

\bibitem[Torquato and Stell(1982)]{torquato1982}
S.~Torquato and G.~Stell.
\newblock Microstructure of two‐phase random media. i. the n‐point
  probability functions.
\newblock \emph{The Journal of Chemical Physics}, 77\penalty0 (4):\penalty0
  2071--2077, 1982.
\newblock \doi{10.1063/1.444011}.

\bibitem[Ulyanov et~al.(2017{\natexlab{a}})Ulyanov, Vedaldi, and
  Lempitsky]{ulyanov2017deep}
Dmitry Ulyanov, Andrea Vedaldi, and Victor Lempitsky.
\newblock Deep image prior.
\newblock \emph{arXiv preprint arXiv:1711.10925}, 2017{\natexlab{a}}.

\bibitem[Ulyanov et~al.(2017{\natexlab{b}})Ulyanov, Vedaldi, and
  Lempitsky]{ulyanov2017improved}
Dmitry Ulyanov, Andrea Vedaldi, and Victor Lempitsky.
\newblock Improved texture networks: Maximizing quality and diversity in
  feed-forward stylization and texture synthesis.
\newblock In \emph{Proc. CVPR}, 2017{\natexlab{b}}.

\bibitem[Vo and Durlofsky(2014)]{vo2014new}
Hai~X Vo and Louis~J Durlofsky.
\newblock A new differentiable parameterization based on principal component
  analysis for the low-dimensional representation of complex geological models.
\newblock \emph{Mathematical Geosciences}, 46\penalty0 (7):\penalty0 775--813,
  2014.

\bibitem[Wu et~al.(1999)Wu, Zhu, and Liu]{wu1999equivalence}
Ying~Nian Wu, Song~Chun Zhu, and Xiuwen Liu.
\newblock Equivalence of julesz and gibbs texture ensembles.
\newblock In \emph{Computer Vision, 1999. The Proceedings of the Seventh IEEE
  International Conference on}, volume~2, pages 1025--1032. IEEE, 1999.

\end{thebibliography}

\appendix
\section{Implementation details}\label{sec:details}

\subsection{Autoencoder} \label{sec:autoencoder}

The architecture of the autoencoder is designed based on the template
provided in~\citep{radford2015unsupervised}: The encoder
$h\colon\tilde{\manX}\to\RR^8$ is a chain of convolutions with leaky
ReLU activations, with $\tanh$ activation in the last layer.
The decoder $d\colon\RR^8\to\tilde{\manX}$ is a chain of transposed convolutions
with ReLU activations, also with $\tanh$ in the final layer.
The code size is $8$. The architecture is detailed in~\Cref{table:autoencoder}.
We train to minimize $\argmin_{d, h} {||x - d(h(x))||^2}$ on patches of
the exemplar image. To reduce overfitting, we data-augment by performing
horizontal and vertical flips on the patches, as well as smoothing by adding
a small amount of Gaussian noise (with $0.05$ standard deviation). We use the
Adam optimizer with default parameters and learning rate $10^{-3}$, and train for $2000$ iterations using
a batch size of $32$. The model takes a few seconds to train using a GTX Titan X.
The decoder is discarded after training and we keep the encoder for the MMD kernel.

\subsection{Generator} \label{sec:generator} 

The generator is also designed based on the template provided
in~\citep{radford2015unsupervised}, but we replace most of the transposed
convolutions with upsampling + convolution (motivated
by~\citep{odena2016deconvolution}), and add an additional convolving layer
before the output. Specifically, the transposed convolutions are replaced by
a $\times 2$ nearest neighbor upsampling followed by a convolution. The
activation in the last layer is $\tanh$. The architecture is detailed
in~\Cref{table:generator}. We use the Adam optimizer with default parameters
and learning rate $10^{-3}$, and train for \num{50000} iterations for both the
$256\times 256$ and $512\times 512$ generators. Using a GTX Titan X, training
takes about $2.5$ and $5$ hours for sizes $256\times 256$ and $512\times
512$, respectively~\footnote{The training is slow in our current
implementation due to the way the patches are being extracted.}. In the
online phase, the generators can synthesize images at the rate of
approximately 150/s and 50/s for sizes $256\times 256$ and $512\times 512$,
respectively.

\begin{table}\centering\small
	\begin{subtable}[t]{.45\textwidth}\centering
        \begin{tabular}[t]{l l}
            \toprule
			State size              & Layer                      \\
			\midrule
			$  1\times 64\times 64$ & \texttt{Conv(4,2,1), BN, lReLU} \\ 
			$ 32\times 32\times 32$ & \texttt{Conv(4,2,1), BN, lReLU} \\ 
			$ 64\times 16\times 16$ & \texttt{Conv(4,2,1), BN, lReLU} \\ 
			$128\times  8\times 8 $ & \texttt{Conv(4,2,1), BN, lReLU} \\ 
			$256\times  4\times 4 $ & \texttt{Conv(4,1,0), Tanh}          \\
			$ 8\times  1\times 1 $ & --                               \\
			\bottomrule
		\end{tabular}
		\caption{Encoder}
    \end{subtable}\hfill
	\begin{subtable}[t]{.45\textwidth}\centering
        \begin{tabular}[t]{l l}
            \toprule
			State size              & Layer                  \\
			\midrule
			$ 8\times  1\times 1 $ & \texttt{ConvT(4,1,0), BN, ReLU} \\ 
			$256\times  4\times 4 $ & \texttt{ConvT(4,2,1), BN, ReLU} \\ 
			$128\times  8\times 8 $ & \texttt{ConvT(4,2,1), BN, ReLU} \\ 
			$ 64\times 16\times 16$ & \texttt{ConvT(4,2,1), BN, ReLU} \\ 
			$ 32\times 32\times 32$ & \texttt{ConvT(4,2,1), Tanh} \\
			$  1\times 64\times 64$ & --     \\ 
			\bottomrule
		\end{tabular}
		\caption{Decoder}
    \end{subtable}
    \caption{Autoencoder architecture\label{table:autoencoder}.
    \texttt{Conv/ConvT}=convolution/transposed convolution, the triplet
    indicates (filter size, stride, padding), \texttt{BN}=batch normalization.}
\end{table}

\begin{table}\centering\small
	\begin{subtable}[t]{.45\textwidth}
        \begin{tabular}[t]{l l}
            \toprule
			State size               & Layer                            \\
			\midrule
			$256\times   1\times 1 $ & \texttt{ConvT(4,1,0), BN, ReLU}  \\ 
			$2048\times  4\times 4 $ & \texttt{UpConv(3,1,1), BN, ReLU} \\ 
			$1024\times  8\times 8 $ & \texttt{UpConv(3,1,1), BN, ReLU} \\ 
			$512\times 16\times 16$  & \texttt{UpConv(3,1,1), BN, ReLU} \\ 
			$256\times  32\times 32$ & \texttt{UpConv(3,1,1), BN, ReLU} \\
			$128\times 64\times 64$  & \texttt{UpConv(3,1,1), BN, ReLU} \\
			$64\times 128\times 128$ & \texttt{UpConv(3,1,1), BN, ReLU} \\
			$64\times 256\times 256$ & \texttt{Conv(3,1,1), Tanh}       \\
			$1\times 256\times 256$  & --                               \\
			\bottomrule
		\end{tabular}
		\caption{$256\times 256$ generator.}
	\end{subtable}\hfill
	\begin{subtable}[t]{.45\textwidth}
        \begin{tabular}[t]{l l}
            \toprule
			State size                & Layer                            \\
			\midrule
			$512\times   1\times 1 $  & \texttt{ConvT(4,1,0), BN, ReLU}  \\ 
			$4096\times  4\times 4 $  & \texttt{UpConv(3,1,1), BN, ReLU} \\ 
			$2048\times  8\times 8 $  & \texttt{UpConv(3,1,1), BN, ReLU} \\ 
			$1024\times 16\times 16$  & \texttt{UpConv(3,1,1), BN, ReLU} \\ 
			$512\times  32\times 32$  & \texttt{UpConv(3,1,1), BN, ReLU} \\
			$256\times 64\times 64$   & \texttt{UpConv(3,1,1), BN, ReLU} \\
			$128\times 128\times 128$ & \texttt{UpConv(3,1,1), BN, ReLU} \\
			$64\times 256\times 256$  & \texttt{UpConv(3,1,1), BN, ReLU} \\
			$64\times 512\times 512$  & \texttt{Conv(3,1,1), Tanh}       \\
			$1\times 512\times 512$   & --                               \\
			\bottomrule
		\end{tabular}
		\caption{$512\times 512$ generator.}
	\end{subtable}
	\caption{Generator architecture\label{table:generator}.
        \texttt{UpConv}=$\times 2$ upsample + convolution, 
        \texttt{ConvT}=transposed convolution, 
        the triplet indicates (filter size, stride, padding), \texttt{BN}=batch normalization.}
\end{table}

\section{Additional results} \label{sec:additional}
\begin{figure}\centering
	\begin{subfigure}{.9\textwidth}\centering
		\includegraphics[width=\textwidth]{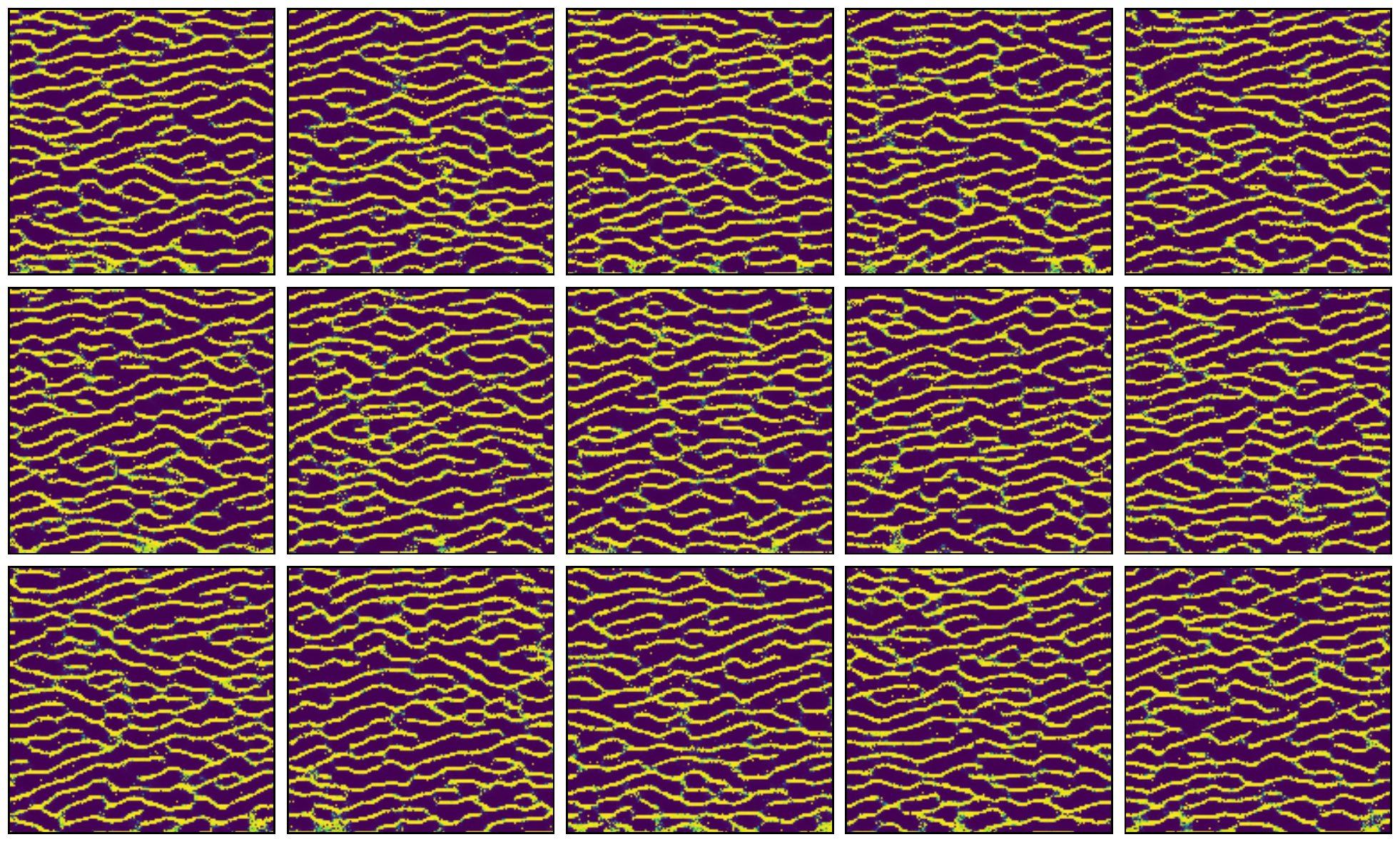}
		\caption{Random realizations ($512\times512$, optimization-based synthesis).}
	\end{subfigure}\vspace{1em}
	
	\begin{subfigure}{.8\textwidth}\centering
		\includegraphics[width=\textwidth]{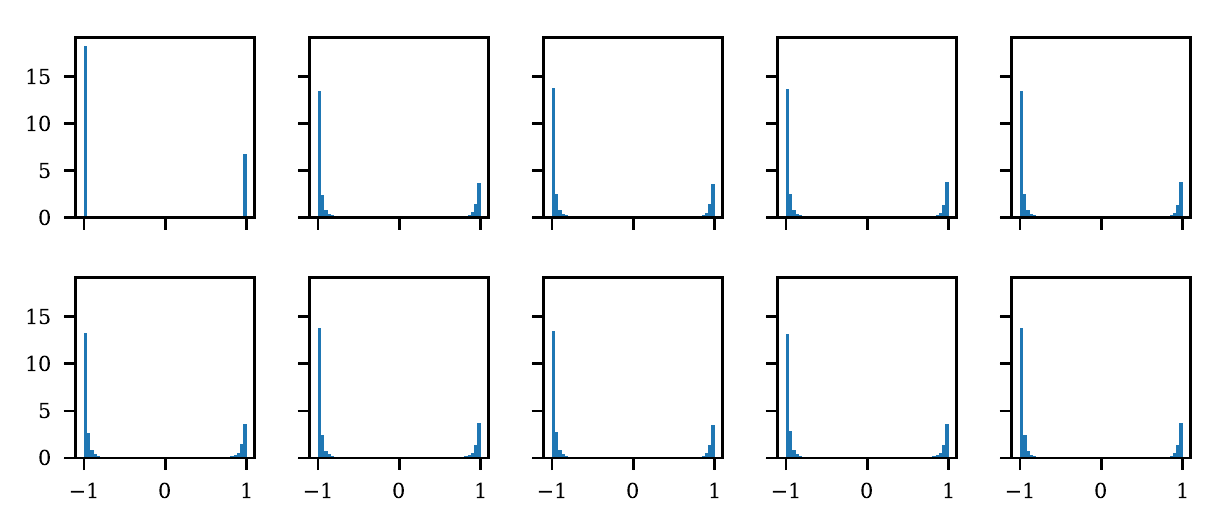}
		\caption{Image histogram of 9 random realizations. The first histogram (top left) corresponds to the exemplar image.}
	\end{subfigure}\vspace{1em}
	
	\begin{subfigure}{.45\textwidth}\centering
		\includegraphics[width=\textwidth]{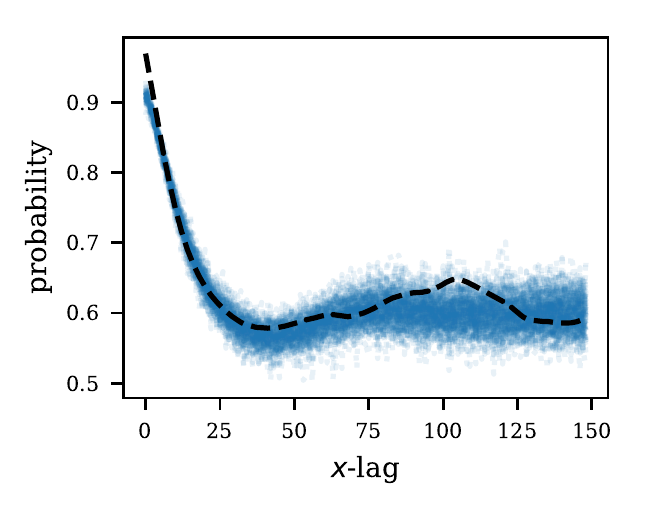}
		\caption{Two-point probability in the $x$ direction of 100 realizations. }
	\end{subfigure}
	\hfill
	\begin{subfigure}{.45\textwidth}\centering
		\includegraphics[width=\textwidth]{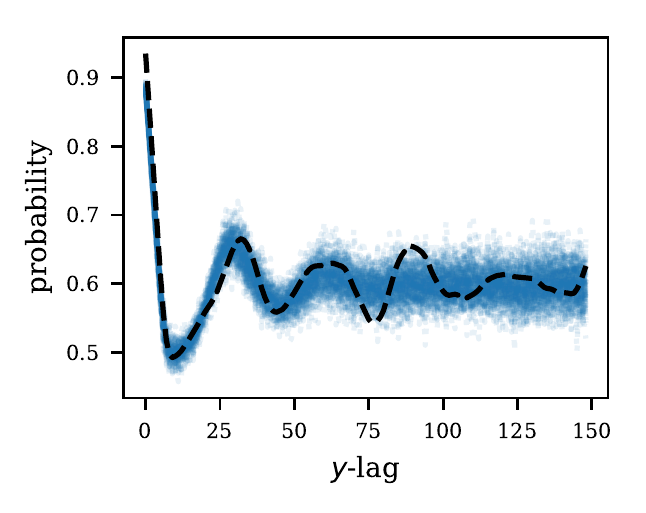}
		\caption{Two-point probability in the $y$ direction of 100 realizations. }
	\end{subfigure}
	\caption{Results for \emph{optimization-based synthesis} of realizations of size $512\times512$ with $k_{\textrm{rq,encoder}}$ kernel. \label{fig:optim512}}
\end{figure}

\begin{figure}\centering
	\begin{subfigure}{.9\textwidth}\centering
		\includegraphics[width=\textwidth]{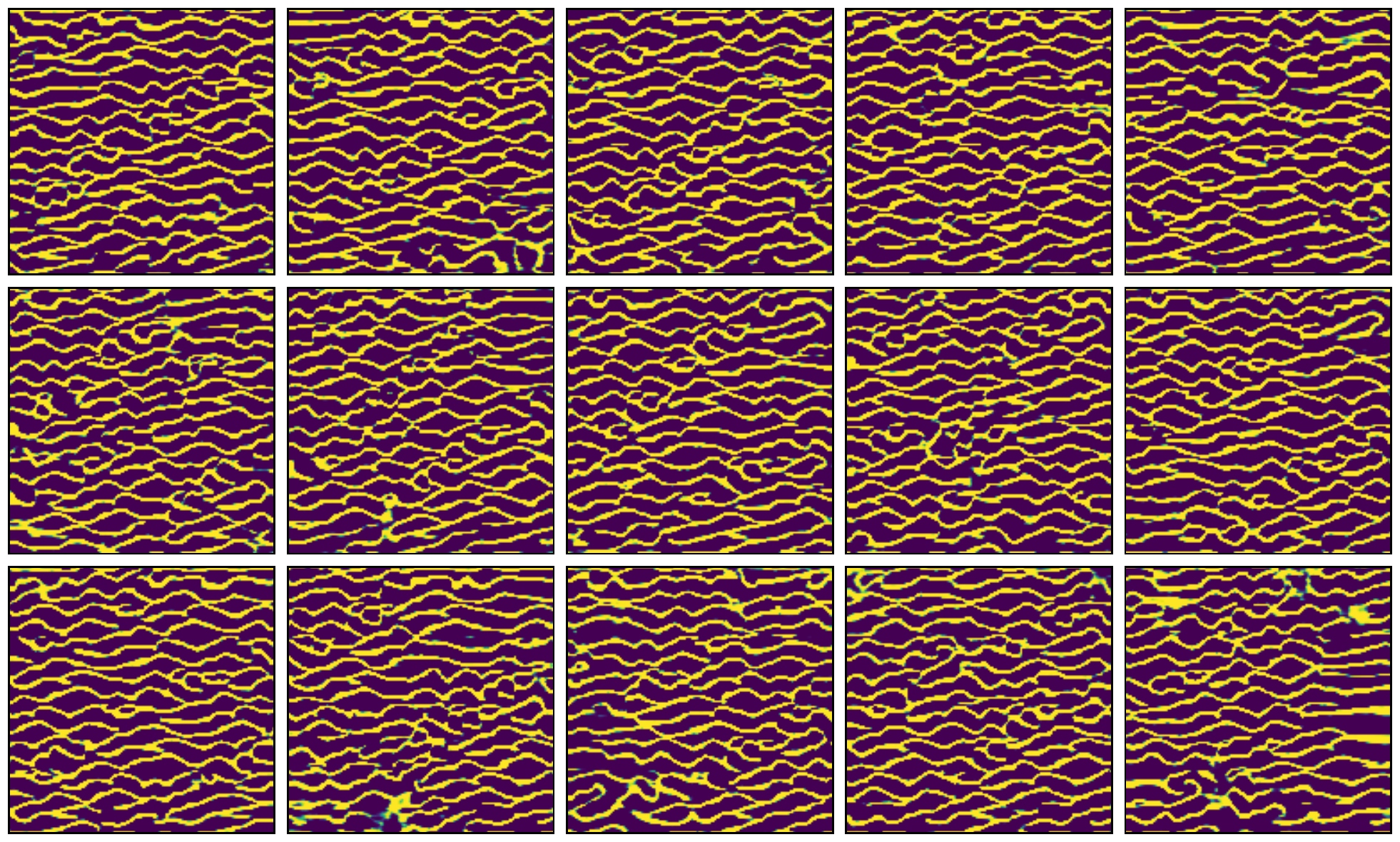}
		\caption{Random realizations ($512\times512$, generated by neural network).}
	\end{subfigure}\vspace{1em}
	
	\begin{subfigure}{.8\textwidth}\centering
		\includegraphics[width=\textwidth]{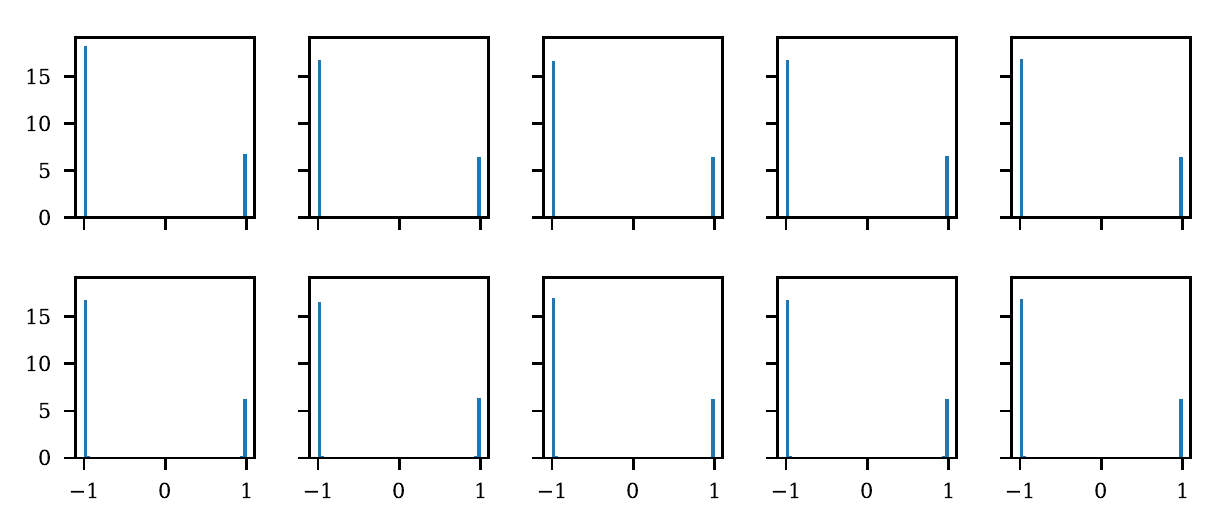}
		\caption{Image histogram of 9 random realizations. The first histogram (top left) corresponds to the exemplar image.}
	\end{subfigure}\vspace{1em}
	
	\begin{subfigure}{.45\textwidth}\centering
		\includegraphics[width=\textwidth]{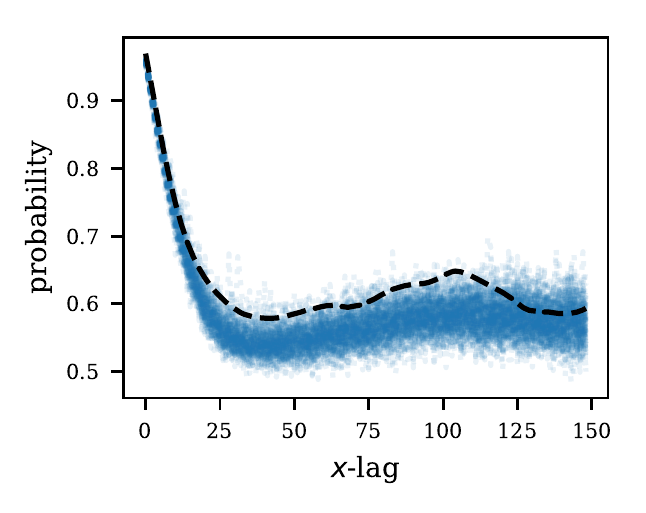}
		\caption{Two-point probability in the $x$ direction of 100 realizations. }
	\end{subfigure}
	\hfill
	\begin{subfigure}{.45\textwidth}\centering
		\includegraphics[width=\textwidth]{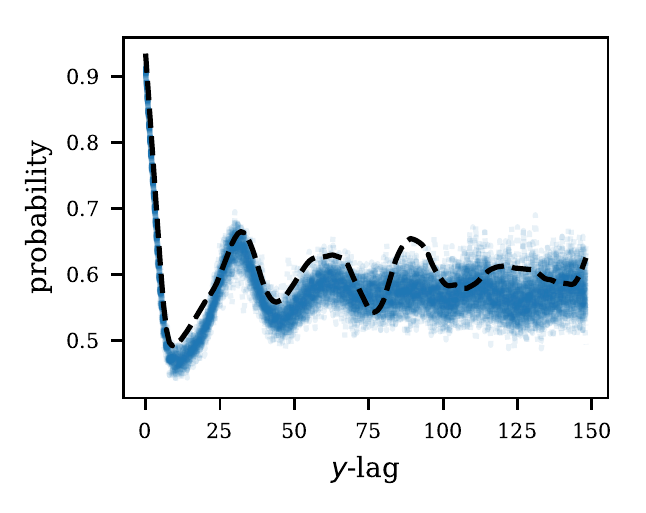}
		\caption{Two-point probability in the $y$ direction of 100 realizations. }
	\end{subfigure}
	\caption{Results for \emph{neural synthesis} of realizations of size $512\times512$ with $k_{\textrm{rq,encoder}}$ kernel.}
\end{figure}


\end{document}